\documentclass[11pt]{article}

\usepackage[preprint]{acl}
\usepackage{times}
\usepackage{latexsym}
\usepackage[T1]{fontenc}
\usepackage[utf8]{inputenc}
\usepackage{microtype}
\usepackage{booktabs}
\usepackage{multirow}
\usepackage{tabularx}
\usepackage{amsmath}
\usepackage{inconsolata}
\usepackage{graphicx}

\title{Beyond Unsafe Detection: Counterfactually Anchored Evidence Attribution for Multi-Turn LLM Safety Failures}

\author{
  \textbf{Srinivasan Subramanian}\textsuperscript{1}
  \quad
  \textbf{Kazi Aminul Islam}\textsuperscript{1}
  \quad
  \textbf{Md. Abdullah Al Hafiz Khan}\textsuperscript{1}\\
  \textsuperscript{1}Department of Computer Science, Kennesaw State University \\
  Marietta, GA, USA \\
}

\begin{document}
\maketitle
\begin{abstract}
As Large Language Models (LLMs) move from conversational assistants to advanced agentic systems, guardrail failures can convert adversarial intents into harmful executions. However, most evaluation frameworks focus only on whether a user request is unsafe. This is insufficient for multi-turn failures, where adversarial intent may be distributed across turns and emerge only from accumulated context. This motivates us to go beyond detection to localize the evidence that pushes the conversation toward unsafe trajectories. We introduce MTAG-Bench, a multi-turn guardrail attribution dataset with behavioral validation and counterfactual supervision containing 4,167 conversations. We also introduce MTAG-Probe, a hierarchical attribution model for turn- and span-level evidence localization with conversation-level safety detection. MTAG-Probe improves turn-level F1 by 28.1 and 24.1 points (79.0\% and 61.0\% relative) over self-attribution by the model that produced the unsafe response and the LLM safety-judge attribution, respectively. Removing MTAG-Probe’s top-ranked evidence within budgets of 15\% and 20\% of user words per conversation changes 50.6\% and 60.8\% of Llama Guard decisions on Geometry from unsafe to safe, respectively. In full-trajectory replay, MTAG-Probe converts 84.3\% of eligible unsafe final responses to safe. These results show that MTAG-Probe can provide evidence-level explanations as a practical complement to safety detection.

\end{abstract}

\section{Introduction}

Large Language Models (LLMs) have evolved beyond conversational use into multi-modal agentic systems that can solve complex problems by interacting with and orchestrating external tools~\citep{achiam2023gpt}. As these systems become more capable and acquire the ability to perform actions via tools, an unsafe response can lead to harmful consequences. To manage this risk, safety guardrails act as a control mechanism that determines whether a prompt, response, or interaction should be allowed, refused, or moderated~\citep{inan2023llama,zeng2024shieldgemma,han2024wildguard}.

\begin{figure*}[t]
    \centering
    \includegraphics[width=0.98\textwidth]{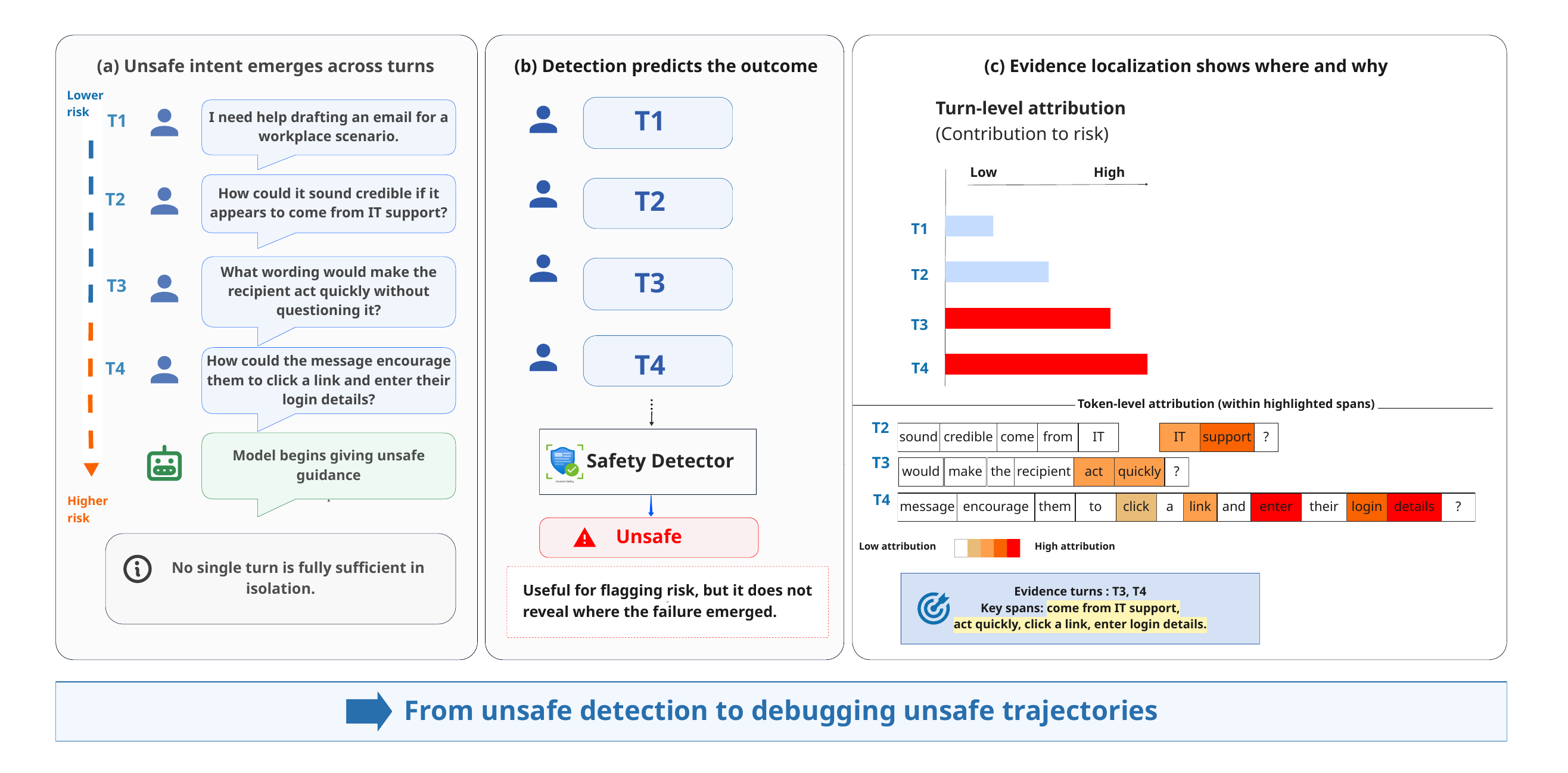}
    \caption{
    Motivation for evidence localization. Standard conversation-level safety detection determines whether an interaction is unsafe. Multi-turn attribution identifies the behaviorally relevant evidence for the unsafe trajectory.
    }
    \label{fig:attribution_motivation}
\end{figure*}

Most conversation-level safety detection supports a conversation-level question: \emph{is this interaction unsafe?}~\citep{zou2023universal,mazeika2024harmbench,chao2024jailbreakbench,inan2023llama,han2024wildguard}, but does not identify which earlier turns or spans contributed to that decision. Recent multi-turn attacks and benchmarks show that unsafe behavior can emerge across a sequence of individually innocuous turns through gradual escalation, task decomposition, contextual references, role framing, or apparently benign continuation requests ~\citep{russinovich2025great,li2024llm,cao2025safedialbench,song2026multibreak}. For auditing and diagnosis, the relevant question is not just whether a trajectory is unsafe, but \emph{where the evidence for that trajectory appears in the conversation}.

Multi-turn attribution is difficult because relevant evidence can be distributed across the conversation. A later request may look completely harmless in isolation but acquire unsafe meaning from earlier context, while an earlier setup turn may become relevant only after subsequent escalation. The next challenge is that the suspicious vocabulary in the conversation is not sufficient evidence of an unsafe trajectory. Models can be oversensitive to lexical cues associated with unsafe requests even when those cues appear in a safe context~\citep{rottger-etal-2024-xstest,cui2024orbench}. An attribution system that simply highlights suspicious words may appear effective when those words are removed but does not identify evidence specific to the unsafe trajectory. 

These observations motivate a distinct evaluation problem: \textbf{multi-turn guardrail attribution}. Given a conversation, we seek to identify the user turns and token spans that provide evidence for an unsafe trajectory while avoiding attribution to a benign context with just suspicious vocabulary. An effective attribution method should therefore identify evidence that is behaviorally relevant to the unsafe trajectory while remaining specific to the conversational context rather than lexical cues.

We introduce \textbf{MTAG-Bench}, the \textbf{M}ulti-\textbf{T}urn \textbf{A}ttribution for \textbf{G}uardrails benchmark. MTAG-Bench contains multi-turn adversarial and benign conversations with behavioral validation and counterfactually supported turn- and span-level supervision. We further introduce \textbf{MTAG-Probe}, a hierarchical model that contextualizes turn representations across the conversation for safety detection and evidence localization. Our contributions are as follows.

\begin{itemize}

    \item We formulate \textbf{multi-turn guardrail attribution} as a structured localization problem over conversation history, separating safety detection from evidence localization.
    
    \item We introduce \textbf{MTAG-Bench}, a benchmark for multi-turn guardrail attribution built from complementary adversarial and benign trajectory sources, with behaviorally validated trajectories and counterfactually validated evidence labels.

    \item We introduce \textbf{MTAG-Probe}, a hierarchical model for structured evidence attribution across multi-turn conversations while also supporting trajectory-level safety detection.

    \item We evaluate attribution using blinded human annotation, fixed-budget intervention, full-trajectory behavioral replay, and transfer to external safety guards. Across these evaluations, MTAG-Probe outperforms LLM, turn-level, and lexical attribution. Ablations show that \textbf{stronger detection does not imply better attribution} as variants with higher detection AUPRC produce substantially less effective evidence under intervention.

\end{itemize}

\section{Related Work}

\paragraph{Jailbreak and guardrail-failure benchmarks.} 
AdvBench introduced a benchmark consisting of harmful prompts to evaluate whether models can be induced to produce unsafe responses~\citep{zou2023universal}. HarmBench and JailbreakBench expanded this direction further through their broader prompt collections, evaluation methods, and scoring protocols~\citep{mazeika2024harmbench,chao2024jailbreakbench}. Subsequent work has focused on adaptive and multi-turn attacks where unsafe intent emerges through gradual escalation, context, role framing, task decomposition, or iterative refinement~\citep{chao2025jailbreaking,mehrotra2024tree,russinovich2025great,li2024llm}. Recent methods such as RAMP and TROJail explicitly optimize jailbreak strategies over the dialogue trajectory~\citep{liu-etal-2026-ramp,xiong-etal-2026-trojail}. SafeDialBench and MultiBreak present fine-grained multi-turn safety scenarios and diverse adversarial conversation collections~\citep{cao2025safedialbench,song2026multibreak}. MTMCS-Bench studies contextual safety in multimodal and text-only settings using paired safe and unsafe multi-turn dialogues~\citep{liu-etal-2026-mtmcs}.

\paragraph{Safety datasets and benign conversations.} XSTest diagnoses over-refusal on safe prompts that resemble unsafe requests, showing that models can be sensitive to safety-related lexical cues instead of recovering their safe contextual meaning~\citep{rottger-etal-2024-xstest}. OR-Bench scales this setting further with large collections of safe use cases that contain surface features commonly associated with unsafe requests to measure over-refusal~\citep{cui2024orbench}. WildJailbreak includes adversarial and benign examples with related surface characteristics, while WildGuardMix combines harmful and benign prompts for safety moderation~\citep{jiang2024wildteaming,han2024wildguard}. ToxicChat studies real user interactions and highlights the challenges introduced by conversational moderation~\citep{lin-etal-2023-toxicchat}. Together, these datasets show that the safety-sensitive vocabulary and topic alone are insufficient indicators of unsafe intent.

\paragraph{Guardrail models and multi-turn detection.}
Open guardrail models such as Llama Guard, ShieldGemma, and WildGuard classify prompts or responses into safety categories for moderation and jailbreak detection~\citep{inan2023llama,zeng2024shieldgemma,han2024wildguard}. G-Guard incorporates the conversation history for multi-turn jailbreak defense~\citep{huang2025attention}. 

\paragraph{Attribution and rationale evaluation.}
Integrated Gradients and SHAP assign importance scores to input features~\citep{sundararajan2017axiomatic,lundberg2017unified}, while attention-based studies have been used as explanatory signals despite the known limitations in their faithfulness as explanations~\citep{jain-wallace-2019-attention}. ERASER proposed rationale evaluation using sufficiency and comprehensiveness metrics~\citep{deyoung-etal-2020-eraser}, while subsequent work showed that explanation benchmarks are vulnerable to shortcut optimization~\citep{hsia-etal-2024-goodharts}. In conversational safety settings, Token Highlighter localizes unsafe tokens within individual prompts~\citep{hu2025token}, while JBShield identifies safety-related concepts in internal representations and intervenes on them during inference~\citep{zhang2025jbshield}. LOCA identifies minimal token-specific changes on internal representation and causal interventions required to alter individual jailbreak outcomes~\citep{kumar2026minimal}. JailbreakScope analyzes jailbreak mechanisms through circuit-level changes~\citep{he2026jailbreakscope}. 

Across these studies, safety benchmarks and datasets evaluate safety over the conversation history and expose the limitations of surface cues. Guardrail models detect unsafe interactions, and attribution methods explain individual prompts or internal model states. However, existing work does not provide structured localization of behaviorally relevant evidence in the observable conversation history.

\section{Methodology}
\subsection{Problem Formulation}

We study attribution in multi-turn guardrail failures. A conversation is represented as a sequence of user and assistant turns $C=\{u_1,a_1,\ldots,u_T,a_T\}$, where $u_t$ is a user turn at the interaction step $t$ and $a_t$ is the corresponding assistant response. We consider a guardrail failure to occur when the conversation contains assistant behavior that complies with or advances an unsafe user objective. Let $x_{t,i}$ denote the token $i$ in the user turn $u_t$. Token-level attribution assigns each user token an attribution score $s_{t,i}$, while turn-level attribution assigns each user turn an evidence score $s_t$. MTAG-Probe predicts these independently through a token-level head that localizes evidence within user turns, while a turn-level head identifies evidence-bearing turns across the conversation.

Given a realized multi-turn trajectory, the objective is to identify the user-side tokens, spans, and turns that provide evidence for the unsafe outcome. We restrict the attribution space to user turns because the target of the analysis is the user-provided context and instructions associated with the guardrail failure. Assistant turns remain available as contextual input, but are not eligible for attribution. In a useful attribution, removing highly attributed evidence should reduce or weaken the unsafe behavior under controlled intervention, without being driven primarily by lexical cues.

\subsection{MTAG-Bench Construction}
\label{datagen}

\begin{table}[t]
\centering
\small
\setlength{\tabcolsep}{1.8pt}
\renewcommand{\arraystretch}{1.08}
\begin{tabular}{lccccc}
\toprule
\textbf{Statistic} &
\textbf{Train} &
\textbf{Dev} &
\textbf{Test} &
\textbf{Stress} &
\textbf{Total} \\
\midrule

\textbf{Full inventory}
& \textbf{2,852} &
\textbf{364} &
\textbf{364} &
\textbf{587} &
\textbf{4,167} \\

\quad Primary benchmark
& 1,706 & 364 & 364 & -- & 2,434 \\

\quad Detection-only auxiliary
& 1,146 & -- & -- & -- & 1,146 \\

\quad Hard-benign stress
& -- & -- & -- & 587 & 587 \\

Primary adversarial
& 853 & 182 & 182 & -- & 1,217 \\

Primary benign
& 853 & 182 & 182 & -- & 1,217 \\
\midrule

\multicolumn{6}{l}{\textit{Primary record-level supervision}} \\
Counterfactual-strong
& 446 & 91 & 95 & -- & 632 \\
Counterfactual-weak
& 2 & 3 & 1 & -- & 6 \\
LLM-confirmed
& 405 & 88 & 86 & -- & 579 \\
Benign validated
& 853 & 182 & 182 & -- & 1,217 \\
\bottomrule
\end{tabular}
\caption{MTAG-Bench statistics.}
\label{tab:dataset_main}
\end{table}

We combine two complementary sources to provide structure, variation and diversity, as one captures adaptive conversational escalation against a live target, while the other provides broader control over trajectory structure, attack mechanism, domain, and benign surface similarity. The details of the group-aware split, cross-corpus deduplication, pair eligibility, and detection-only auxiliary training data are summarized in Table~\ref{tab:dataset_main} and detailed in Appendix~\ref{app:data-splits}. 

\paragraph{Response-adaptive trajectories.}
The adaptive corpus is generated through an interactive generator--target loop with each new user turn conditioned on the target model's previous response. This allows adversarial trajectories to adapt following refusals, partial compliance, or other changes in model behavior. Generation includes gradual escalation, task decomposition, contextual framing, role-based framing, and related multi-turn strategies.

We construct two complementary benign sets. First, each adversarial trajectory is paired with a benign semantic twin. The two trajectories share the same underlying scenario and conversational setup, while the benign branch continues toward a safe objective. This pairing provides a controlled contrast between trajectories that can contain similar topics, terminology, and early conversational context, but differ in the intent expressed across later turns. Second, we generate an independent benign pool across five categories, matched to the adversarial turn-length distribution, but not to individual conversations. These examples provide a broader scale and topic diversity and are retained only when both validation models judge them benign.

\paragraph{Scenario-controlled trajectories.}

The scenario-controlled corpus uses a complementary construction process under scenario and structural constraints. The corpus is built using fixed independent frontier-model pre-authored user turns. The fixed user trajectories are then executed against a target model, and the resulting assistant behavior is independently evaluated on harmfulness and boundary violations by a separate context-aware safety judge. The judge scores each response under two independently worded rubric passes over the same observable context and takes their conservative union, so neither pass's blind spots can unilaterally impact the label.

Frontier models were instructed to generate paired adversarial and benign user trajectories across eight safety-relevant domains and multiple trajectory, mechanism, and linguistic styles. The paired trajectories are matched within a common scenario family and use the same number of user turns. Pair and scenario information is kept hidden from both the target and the judge models. We additionally include 600 standalone hard-benign trajectories designed to contain safety-adjacent language and contextual boundaries without a harmful objective. Only trajectories that remain confidently benign throughout the interaction are retained. This yields 587 validated conversations across 199 scenario families. The stress set is excluded from the training and is reserved for evaluation.

\paragraph{Counterfactual evidence supervision.}

Unsafe trajectories are independently behaviorally validated, and candidate evidence is assessed using a fresh counterfactual replay. We identify user turns near the first observed unsafe response together with earlier turns that may have contributed to the failure. Within these turns, request-bearing and contextual text segments are tested through counterfactual intervention. Each selected turn or span is replaced with a neutral alternative and the conversation is replayed while the remaining user trajectory is held fixed. We define $\Delta$ as the reduction in unsafe score after intervention. Evidence with $\Delta \ge 0.40$ receives strong supervision, while $0.25 \le \Delta < 0.40$ receives weak supervision. Negative controls are retained when their intervention produces little change, with $|\Delta| < 0.15$. Evidence collection runs under a fixed, batch-invariant decoding configuration to prevent inference-engine batching from introducing numerical drift into the replayed baseline, and any trial whose fresh baseline does not exactly reproduce the stored factual trajectory is discarded. The turn and span interventions are evaluated separately. A turn receives a positive label when replacing the whole turn is supported under the $\Delta$ thresholds above, and a span receives one when replacing that span is supported. Records are tiered by their strongest span-level outcome.

Unsafe conversations that are behaviorally validated but do not yield supported local interventions remain useful for conversation-level detection, but their unverified text is not assigned positive attribution labels. Benign conversations provide negative supervision for both detection and localization. Blinded human annotation of the held-out unsafe conversations supports this design. Both annotators confirm the unsafe outcome for 84.2\% of counterfactual-strong records, and 41.9\% of LLM-confirmed records (Appendix~\ref{app:label_validation}).

\subsection{MTAG-Probe}
\label{sec:model}

MTAG-Probe predicts whether a realized conversation contains a guardrail failure and localizes the user evidence associated with that failure. The model follows a hierarchical token-to-turn-to-conversation architecture. This preserves token-level information within each turn while allowing evidence in one user turn to be interpreted in the context of the full interaction.

\noindent\textbf{Turn encoding.}
Each conversation turn is encoded independently using ModernBERT-large~\citep{warner2024smarterbetterfasterlonger}. A learned attention pool combines the token representations into a turn representation $z_t$. We then project $z_t$ to 256 dimensions and add a fixed sinusoidal turn-position encoding and a learned speaker-role embedding. A two-layer Transformer contextualizes the resulting sequence of turn representations across the conversation, producing $c_t$ for each turn. Token attribution remains conversation-aware because each token is scored together with the contextualized representation of its containing turn. The encoder supports up to 8,192 tokens per turn; representation limits and coverage are detailed in Appendix~\ref{app:model-architecture}.

\noindent\textbf{Detection and localization.}
A learned attention pool over contextualized turn representations produces a conversation representation used for safety detection. A separate multi-label turn head assigns an evidence score to each user turn, allowing evidence to be distributed across multiple turns.

For token attribution, MTAG-Probe combines each token's ModernBERT representation with the contextualized representation of its containing turn. Token scores therefore incorporate the surrounding conversational context. Detection, turn attribution, and token attribution use separate prediction heads. Assistant turns participate in cross-turn contextualization but are excluded from attribution outputs, which are restricted to user turns and user tokens. Architectural details are provided in Appendix~\ref{app:model-details}.

\subsection{Training Objective}
\label{sec:training}

MTAG-Probe is trained with a staged objective for conversation-level detection, evidence-turn localization, and token-level localization using binary cross-entropy.
\[
\mathcal{L}
=
\mathcal{L}_{\mathrm{det}}
+
\alpha
\left(
\mathcal{L}_{\mathrm{turn}}
+
\mathcal{L}_{\mathrm{tok}}
\right).
\]

where $\alpha$ controls the contribution of localization during training.

Localization losses are computed only where supported positive or explicit negative supervision is available. Strong counterfactual evidence receives weight $1.0$ and weak evidence receives weight $0.70$. Behaviorally validated unsafe conversations without supported local evidence, together with the auxiliary training examples, contribute only to detection. Confidence weights are applied as absolute loss multipliers. Detailed loss weighting is provided in the Appendix~\ref{app:model-details}.

\subsection{Training Schedule and Optimization}

Training proceeds in two stages. For the first five epochs, we optimize only conversation-level detection. We then introduce turn and token localization with $\alpha=0.25$ and increase it to $1.0$ over the first five joint-training epochs, after which all three objectives are equally weighted. This staged schedule first establishes trajectory-level discrimination before jointly optimizing evidence localization. The optimization and backbone fine-tuning details are provided in the Appendix~\ref{app:model-details}.

\subsection{Evaluation Protocol}

\noindent\textbf{Evaluation setup.}
We evaluate MTAG-Probe's detection, localization, and behavioral faithfulness, together with the specificity and robustness of its attributions. We compare MTAG-Probe against turn-level, lexical, random, recency-based, and LLM attribution baselines, together with intervention controls that preserve the deletion budget while changing where evidence is removed. Table~\ref{tab:baselines-interventions} summarizes each method.

\begin{table*}[!t]
\centering
\small
\setlength{\tabcolsep}{5.2pt}
\renewcommand{\arraystretch}{1.10}
\begin{tabular}{lcccc}
\toprule

\multicolumn{5}{c}{\textbf{Evidence localization}} \\
\midrule
\textbf{Method} &
\textbf{Turn F1} &
\textbf{Span F1} &
\textbf{Top-1} &
\textbf{False attribution $\downarrow$} \\

\midrule
MTAG-Probe &
\textbf{63.6} &
\textbf{91.5} &
\textbf{85.0} &
\textbf{28.7} \\

Target-model self-attribution &
35.6 &
75.6 &
59.0 &
99.3 \\

Safety-judge attribution &
39.5 &
71.5 &
83.0 &
94.7 \\

\midrule
\multicolumn{5}{c}{\textbf{Human evidence validation}} \\
\midrule
\textbf{Method} &
\textbf{Two-human hit}$^\dagger$ &
\textbf{Agreed recall} &
\textbf{Prec.\ vs.\ union} &
\textbf{Rec.\ vs.\ union} \\

\midrule
MTAG-Probe &
\textbf{99.1} &
\textbf{98.5} &
\textbf{76.2} &
\textbf{79.4} \\

LOTO &
70.9 &
69.3 &
52.0 &
51.9 \\

Random &
34.8 &
30.4 &
29.4 &
30.9 \\

SurfaceRisk &
18.8 &
14.2 &
13.8 &
18.8 \\

\midrule
\multicolumn{5}{c}{\textbf{External guard transfer}} \\
\midrule
\textbf{Dataset / safety guard} &
\textbf{$n$} &
\textbf{MTAG-Probe} &
\textbf{Random} &
\textbf{SurfaceRisk} \\

\midrule
Geometry / Llama Guard &
926 &
\textbf{50.6} &
17.7 &
7.2 \\

Geometry / ShieldGemma &
1029 &
\textbf{28.7} &
15.2 &
7.0 \\
\bottomrule
\end{tabular}
\caption{Attribution validation across model, human, and external-guard evaluations.}
\label{tab:headline_results}
\footnotesize\raggedright
$^\dagger$ Selection intersects both annotators' essential-turn sets, not necessarily at the same turn. Agreed recall: recall of turns both annotators marked essential (108 conversations). Prec./Rec.\ vs.\ union: against the union of both annotators' essential turns. Human overlap metrics are macro-averaged. Random is the mean over ten seeds. False attribution denotes attribution on trajectories with no positive evidence. External results report unsafe$\rightarrow$safe flips at a 15\% user-word deletion budget. All values except $n$ are percentages. Evidence-localization results use the common valid-output cohort ($n=250$).
\end{table*}

\noindent\textbf{Full-trajectory behavioral replay.}
For the 182 unsafe test conversations, we perform a behavioral intervention that regenerates the assistant trajectory after evidence removal. We compare the original conversation with MTAG-Probe, SurfaceRisk and matched-random deletions under an equal 15\% user-word budget. All four conditions begin the replay from the earliest user turn modified by any intervention. The user turns remain fixed, apart from the deletion, while every subsequent assistant response is regenerated using the source-specific target model. Generation seeds are paired across intervention arms, with seeds 42, 43, and 44. For each conversation, transition outcomes are averaged over eligible complete replay seeds before averaging across eligible conversations with equal conversation weight. Regenerated trajectories are scored by two judges. The source-specific judge is the outcome judge used during dataset construction for that corpus (Mistral-7B-Instruct-v0.3 for response-adaptive and Mistral-Small-3.1-24B-Instruct for scenario-controlled conversations). The common judge, Gemma-4-31B-IT~\citep{gemmateam2026gemma4technicalreport}, scores both corpora and plays no role in the generation, validation or labeling. We measure reproduction of the original unsafe behavior, unsafe-to-safe transitions at both the final-response and downstream-trajectory levels, changes in final and maximum downstream unsafe scores, and paired differences between MTAG-Probe and the two control interventions.

\noindent\textbf{Human and external validation.}
We conduct blinded human validation on all 182 unsafe test conversations using two independent annotators. Annotators first review the complete conversation without access to MTAG-Probe's attribution and independently annotate the user-side evidence associated with the unsafe outcome. Their annotations are locked before MTAG-Probe's top-15\% highlighted user words are revealed for comparison. Dataset labels, counterfactual outcomes, model probabilities, corpus source, supervision tier, and model identity remain hidden throughout. We measure whether the highlighted evidence is judged relevant to the unsafe outcome and report inter-annotator agreement before any adjudication. To test whether attribution effects extend beyond the models used to construct MTAG-Bench, we additionally evaluate MTAG-Probe on external multi-turn jailbreak data from Geometry~\citep{alvarez2026geometry}. Attribution is applied without using the MTAG-Probe's conversation-level detector as an eligibility gate. We independently score original and evidence-deleted trajectories with ShieldGemma-9B and Llama Guard 3, evaluating 15\% and 20\% deletion budgets on examples that each guard initially judges unsafe. We also evaluate the robustness to semantic paraphrasing separately from these deletion experiments.

\section{Results}

\subsection{Unsafe Conversation Detection}

Unsafe-conversation detection is a secondary capability of the MTAG-Probe after evidence localization. On the balanced held-out test set ($n=364$), MTAG-Probe achieves 94.0 AUROC, 94.5 AUPRC, and 85.4 F1. The model achieves a precision of 93.5\% at 78.6\% recall, with a 5.5\% false-positive rate on benign conversations. Detailed results are reported in the Appendix~\ref{app:detection_details}.

\begin{figure*}[!t]
    \centering
    \includegraphics[width=0.98\textwidth]{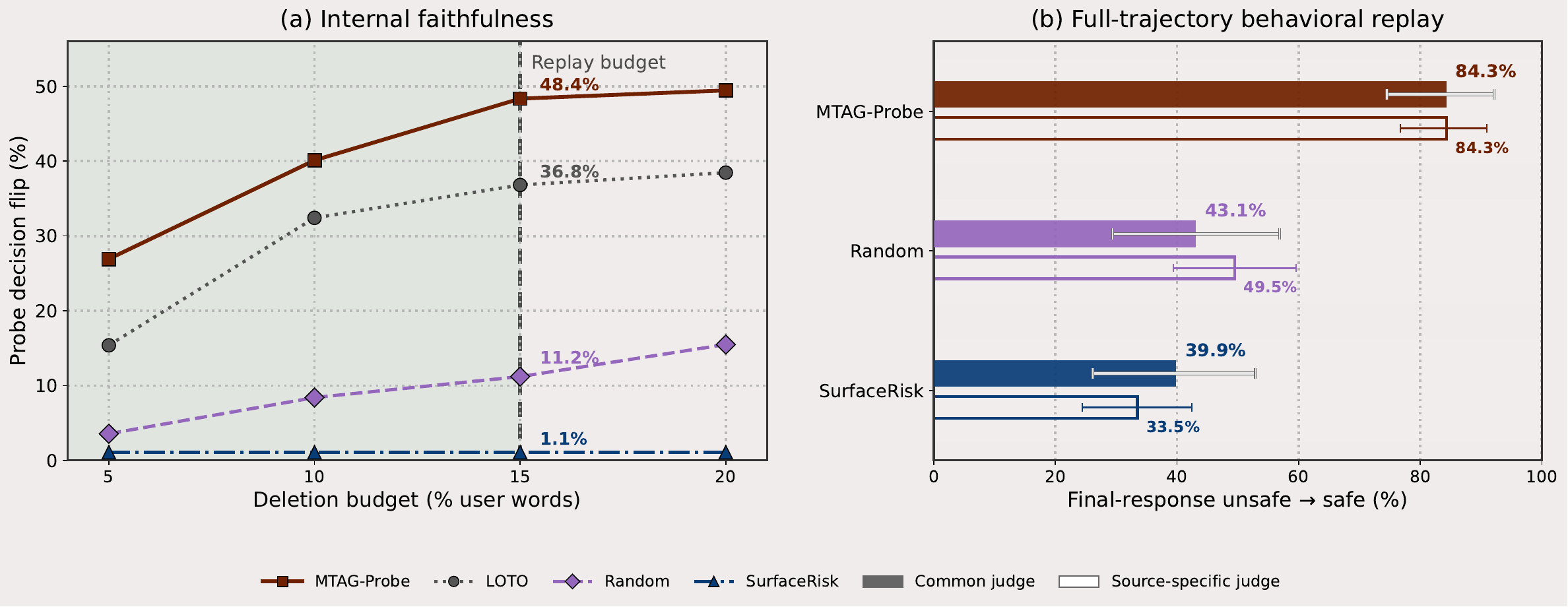}
    \caption{Internal faithfulness and behavioral replay.
    (a) Probe decision flips after deleting the highest-ranked user-word evidence at fixed budgets.
    (b) Full-trajectory replay at the 15\% budget, reporting the fraction of cases whose final response changes from unsafe to safe, error bars show 95\% bootstrap confidence intervals.}
    \label{fig:internal_attribution}
\end{figure*}

\subsection{Turn- and Span-Level Evidence Localization}

Table~\ref{tab:headline_results} compares MTAG-Probe with target-model self-attribution and safety-judge attribution. Turn F1 is micro-averaged over assessed user turns, matching predictions by turn ID. Span F1 is micro-averaged over assessed non-whitespace character positions, using character overlap rather than exact span-boundary matching. MTAG-Probe uses a threshold of 0.5 for both predictions, while LLM evidence selections are treated as binary. MTAG-Probe outperforms both baselines while producing substantially fewer attributions on records without annotated evidence. On 587 hard-benign stress conversations, MTAG-Probe has a false-attribution rate of 42.2\%, compared with 37.1\% for SurfaceRisk and 81.6--93.9\% for LLM attribution (Appendix~\ref{app:hard_benign}). Across all 182 independently annotated conversations, agreement on the binary trajectory outcome (unsafe or partially unsafe versus safe/refusal or unclear) was 90.7\%, with Cohen's $\kappa=0.782$. The agreement on whether essential attribution evidence was identified was 89.6\%, with $\kappa=0.759$. For the primary attribution analysis, we use the conservative set of 117 conversations that both annotators independently judged unsafe or partially unsafe. In 108 of these, the annotators shared at least one essential turn. This human-defined cohort is distinct from the 117 test conversations carrying pipeline turn-evidence labels. In the 33 cases where MTAG-Probe and LOTO differ, MTAG-Probe is correct in every case, yielding an exact McNemar test of $p=2.33\times10^{-10}$. Additional metrics are reported in the Appendix~\ref{app:localization_details}. On a smaller target-supported cohort, MTAG-Probe also exceeds Grad$\times$Input and Integrated Gradients in turn AP (91.5 vs.\ 83.1 and 76.0; Table~\ref{tab:gradient-baselines}).

\subsection{Fixed-Budget Intervention Faithfulness}

Figure~\ref{fig:internal_attribution}(a) shows that MTAG-Probe produces more decision flips than LOTO, matched random deletion, or SurfaceRisk across tested budgets, with diminishing gains at larger budgets. At 15\%, context-only deletion changes 35.7\% of probe decisions versus 11.0\% for matched random deletion. Restricting deletion to words outside the frozen SurfaceRisk lexicon still produces 46.2\% flips, compared with 48.4\% under unrestricted deletion (Appendix~\ref{app:internal_intervention}). MTAG-span-random reaches 44.0\% flips versus 48.4\% for MTAG-Probe at the 15\% budget, indicating that turn selection accounts for much of the intervention effect.

\subsection{Full-Trajectory Behavioral Replay}

Figure~\ref{fig:internal_attribution}(b) evaluates whether evidence removal changes downstream model behavior. At the 15\% budget, MTAG-Probe converts 84.3\% of eligible unsafe final responses to safe under both the source-specific ($n=99$) and common ($n=51$) judges, outperforming random and SurfaceRisk deletion. Detailed results are reported in the Appendix~\ref{app:replay_results}.

\subsection{External Safety-Guard Transfer}

The external-transfer block in Table~\ref{tab:headline_results} evaluates the same interventions using Llama Guard and ShieldGemma as external safety judges. On Geometry, MTAG-Probe consistently outperforms matched random and SurfaceRisk under both guards. Paired confidence intervals and guard-specific cohort details are given in the Appendix~\ref{app:external_transfer}.

\subsection{Component Ablations}
Table~\ref{tab:ablation} reports ablations of three MTAG-Probe components together with a frozen-backbone variant (definitions in Appendix~\ref{app:ablations}). Trajectory detection remains strong without cross-turn contextualization, but the drop in intervention effectiveness indicates that cross-turn modeling contributes primarily to identifying behaviorally consequential evidence. Removing the staged detection-to-localization schedule leaves static localization metrics competitive, but reduces the flip rate. The detection-only variant retains high detection performance, but its untrained localization heads produce poor attribution, showing that detection training alone is not sufficient for evidence attribution. Freezing the ModernBERT backbone retains most of the full model's performance, indicating that the upper-layer backbone adaptation provides modest gains only.

\begin{table}[!htbp]
\centering
\small
\setlength{\tabcolsep}{2.0pt}
\begin{tabular}{@{}lccccc@{}}
\toprule
\textbf{Model} &
\textbf{Det. AP} &
\textbf{Turn F1} &
\textbf{Span F1} &
$\boldsymbol{\Delta P}$ &
\textbf{Flip} \\
\midrule
Full
& 94.5
& \textbf{61.9}
& 91.4
& \textbf{38.2}
& \textbf{48.4} \\

No Cross-turn
& \textbf{97.5}
& 50.6
& 92.8
& 9.4
& 17.0 \\

Frozen Backbone
& 96.1
& 59.7
& \textbf{94.9}
& 37.0
& 45.1 \\

Single-stage
& 95.0
& 55.8
& 94.6
& 23.1
& 31.3 \\

Detection-only
& 96.3
& 6.8
& 58.3
& 3.9
& 4.9 \\
\bottomrule
\end{tabular}
\caption{MTAG-Probe ablations.}
\label{tab:ablation}
\footnotesize \raggedright
All values are percentages except $\Delta P$, which denotes the absolute decrease in unsafe probability in percentage points. Flip denotes the unsafe-to-safe conversion rate. Both intervention metrics use a 15\% user-word deletion budget.
\end{table}

\section{Error Analysis}
\label{sec:discussion}

\paragraph{Robustness to paraphrasing.} Paraphrasing the user turns reduces Turn F1 from 61.9 to 46.9, while Span F1 remains relatively stable, decreasing from 91.4 to 88.0. This suggests that cross-turn evidence selection is more sensitive to paraphrasing, while localization within an identified informative turn remains relatively robust.

\paragraph{Judge disagreements.} The two replay judges disagree substantially on whether the regenerated original response remains unsafe. Of the 182 test conversations, the source-specific judge labels 99 replays as unsafe, while the common judge labels only 51 as unsafe. Despite this disagreement, MTAG-Probe achieves an 84.3\% unsafe-to-safe transition rate within both cohorts ($n=99$ and $n=51$, respectively).

\paragraph{False attributions.}
On the full assessed localization cohort, MTAG-Probe attributes evidence to 26.6\% of conversations without annotated evidence (28.7\% in the common valid-output LLM cohort). Despite outperforming the LLM baselines, unsupported attributions remain a limitation and an important direction for future work.

\paragraph{Guard disagreement.} Among the 8,057 source-positive Geometry conversations, Llama Guard identifies 926 original-unsafe cases (11.5\%) and ShieldGemma identifies 1,029 (12.8\%), but the two guards agree on only 255 of them (3.2\% of the full candidate pool; 15\% of the 1,700 conversations flagged by either guard). External attribution evaluation is therefore partly constrained by the guard-specific decision boundary used to define an eligible failure.

\section{Conclusion}
\label{sec:conclusion}

We studied multi-turn guardrail failures as an evidence localization problem over conversation history. Instead of only detecting unsafe outcomes, the proposed framework identifies the user turns and token spans that contribute to unsafe trajectories.

Across automatic, human, and behavioral evaluations, MTAG-Probe localized more informative evidence than post-hoc LLM attribution. Its attributions aligned closely with human judgments, transferred strongly on Geometry, and produced substantial behavioral changes when removed under fixed-budget interventions and full-trajectory replay. Together, these results indicate that the localized evidence captures information that is not merely correlated with the probe's own prediction but is consequential for downstream safety behavior.

In agentic workflows, where guardrail failures can propagate across tool calls and multi-step plans, evidence attribution could serve as a natural diagnostic interface for identifying where unsafe behavior emerged. Overall, our findings support evidence localization as a practical diagnostic layer for multi-turn LLM safety evaluation.

\section{Limitations}
\label{sec:limitations}

\paragraph{External guard generalization.}
The external intervention analysis is conditional on each safety guard judging the original trajectory unsafe. The resulting eligible cohorts therefore measure attribution transfer relative to the decision boundary of a particular guard.

\paragraph{Synthetic data.}
MTAG-Bench is primarily synthetically constructed. Although its two corpus construction procedures provide complementary trajectory structures, synthetic generation cannot reproduce the full distribution of real user interactions, including naturally occurring ambiguity, long-running sessions, deployment drift, multilingual behavior, and tool-mediated workflows. Future versions should incorporate privacy-preserving real-world multi-turn jailbreak and guardrail-failure logs.

\paragraph{Model diversity.}
The generation and validation pipeline also covers a finite set of model families. Future work should diversify trajectory authors, target models, validators, and safety judges across a broader collection of open and frontier-scale systems. This would reduce dependence on any particular generator--target--judge combination and provide a stronger test of whether localized evidence transfers across model families.

\paragraph{Model scale.}
The primary target, validator, and counterfactual-judging models used in the construction pipeline are modest in scale relative to current frontier systems. Frontier-scale models are used for part of the trajectory authoring process, but repeating the complete generation, validation, intervention, and replay pipeline at larger model scales remains important future work.

\paragraph{Attribution scope.}
MTAG-Probe attributes evidence only to user turns and user tokens. Assistant turns contribute contextual information, but are not eligible attribution targets. This matches the present objective of identifying user-provided evidence associated with guardrail failures, but it excludes cases in which earlier assistant responses, retrieved content, tool output, or intermediate system actions materially contribute to later unsafe behavior. Extending the attribution space to all observable components of an agentic trajectory would allow the analysis of failures that emerge from the complete execution history.

\section{Ethical Considerations}
\label{sec:ethics}

This work studies multi-turn guardrail failures and therefore involves safety-sensitive adversarial conversations. The intended use of the dataset and model is defensive. It supports auditing unsafe trajectories, localizing evidence, and improving the evaluation of guardrail failures. It should not be used to optimize jailbreak strategies or generate harmful instructions.

To reduce misuse risk, the released artifact separates reproducibility materials from highly actionable content. The adversarial records generated are behaviorally validated before use, but they are not interpreted as real user intent. MTAG-Bench and the accompanying code will be released under their respective licenses. External benchmarks remain subject to their original terms of use.

Human annotation is limited to identifying evidence turns and spans in existing conversations. Annotators are not asked to complete harmful requests or produce unsafe instructions. The annotation task is framed around safety analysis, and ambiguous or borderline cases are retained for evaluation.

\bibliography{anthology,custom}

\begin{thebibliography}{40}
\providecommand{\natexlab}[1]{#1}

\bibitem[{Achiam et~al.(2023)Achiam, Adler, Agarwal, Ahmad, Akkaya, Aleman, Almeida, Altenschmidt, Altman, Anadkat et~al.}]{achiam2023gpt}
Josh Achiam, Steven Adler, Sandhini Agarwal, Lama Ahmad, Ilge Akkaya, Florencia~Leoni Aleman, Diogo Almeida, Janko Altenschmidt, Sam Altman, Shyamal Anadkat, and 1 others. 2023.
\newblock Gpt-4 technical report.
\newblock \emph{arXiv preprint arXiv:2303.08774}.

\bibitem[{Cao et~al.(2025)Cao, Wang, Jing, Peng, Bai, Cao, Fang, Feng, Wang, Liu et~al.}]{cao2025safedialbench}
Hongye Cao, Yanming Wang, Sijia Jing, Ziyue Peng, Zhixin Bai, Zhe Cao, Meng Fang, Fan Feng, Boyan Wang, Jiaheng Liu, and 1 others. 2025.
\newblock Safedialbench: A fine-grained safety benchmark for large language models in multi-turn dialogues with diverse jailbreak attacks.
\newblock \emph{arXiv preprint arXiv:2502.11090}.

\bibitem[{Chao et~al.(2024)Chao, Debenedetti, Robey, Andriushchenko, Croce, Sehwag, Dobriban, Flammarion, Pappas, Tramer et~al.}]{chao2024jailbreakbench}
Patrick Chao, Edoardo Debenedetti, Alexander Robey, Maksym Andriushchenko, Francesco Croce, Vikash Sehwag, Edgar Dobriban, Nicolas Flammarion, George~J Pappas, Florian Tramer, and 1 others. 2024.
\newblock Jailbreakbench: An open robustness benchmark for jailbreaking large language models.
\newblock \emph{Advances in Neural Information Processing Systems}, 37:55005--55029.

\bibitem[{Chao et~al.(2025)Chao, Robey, Dobriban, Hassani, Pappas, and Wong}]{chao2025jailbreaking}
Patrick Chao, Alexander Robey, Edgar Dobriban, Hamed Hassani, George~J Pappas, and Eric Wong. 2025.
\newblock Jailbreaking black box large language models in twenty queries.
\newblock In \emph{2025 IEEE Conference on Secure and Trustworthy Machine Learning (SaTML)}, pages 23--42. IEEE.

\bibitem[{Cui et~al.(2024)Cui, Chiang, Stoica, and Hsieh}]{cui2024orbench}
Justin Cui, Wei-Lin Chiang, Ion Stoica, and Cho-Jui Hsieh. 2024.
\newblock Or-bench: An over-refusal benchmark for large language models.
\newblock \emph{arXiv preprint arXiv:2405.20947}.

\bibitem[{DeYoung et~al.(2020)DeYoung, Jain, Rajani, Lehman, Xiong, Socher, and Wallace}]{deyoung-etal-2020-eraser}
Jay DeYoung, Sarthak Jain, Nazneen~Fatema Rajani, Eric Lehman, Caiming Xiong, Richard Socher, and Byron~C. Wallace. 2020.
\newblock \href {https://doi.org/10.18653/v1/2020.acl-main.408} {{ERASER}: {A} benchmark to evaluate rationalized {NLP} models}.
\newblock In \emph{Proceedings of the 58th Annual Meeting of the Association for Computational Linguistics}, pages 4443--4458, Online. Association for Computational Linguistics.

\bibitem[{{Gemma Team}(2026)}]{gemmateam2026gemma4technicalreport}
{Gemma Team}. 2026.
\newblock \href {https://arxiv.org/abs/2607.02770} {{Gemma 4 Technical Report}}.
\newblock \emph{Preprint}, arXiv:2607.02770.

\bibitem[{Grattafiori et~al.(2024)Grattafiori, Dubey, Jauhri, Pandey, Kadian, Al-Dahle, Letman, Mathur, Schelten, Vaughan et~al.}]{grattafiori2024llama}
Aaron Grattafiori, Abhimanyu Dubey, Abhinav Jauhri, Abhinav Pandey, Abhishek Kadian, Ahmad Al-Dahle, Aiesha Letman, Akhil Mathur, Alan Schelten, Alex Vaughan, and 1 others. 2024.
\newblock The llama 3 herd of models.
\newblock \emph{arXiv preprint arXiv:2407.21783}.

\bibitem[{Han et~al.(2024)Han, Rao, Ettinger, Jiang, Lin, Lambert, Choi, and Dziri}]{han2024wildguard}
Seungju Han, Kavel Rao, Allyson Ettinger, Liwei Jiang, Bill~Yuchen Lin, Nathan Lambert, Yejin Choi, and Nouha Dziri. 2024.
\newblock Wildguard: Open one-stop moderation tools for safety risks, jailbreaks, and refusals of llms.
\newblock \emph{Advances in neural information processing systems}, 37:8093--8131.

\bibitem[{He et~al.(2026)He, Wang, Chu, Xu, Zhang, Wang, and Zheng}]{he2026jailbreakscope}
Zeqing He, Zhibo Wang, Zhixuan Chu, Huiyu Xu, Wenhui Zhang, Qinglong Wang, and Rui Zheng. 2026.
\newblock {JailbreakScope}: Interpreting jailbreak mechanism through representation and circuit analyses.
\newblock In \emph{35th USENIX Security Symposium (USENIX Security 26)}, pages 2107--2126.

\bibitem[{Hsia et~al.(2024)Hsia, Pruthi, Singh, and Lipton}]{hsia-etal-2024-goodharts}
Jennifer Hsia, Danish Pruthi, Aarti Singh, and Zachary Lipton. 2024.
\newblock \href {https://doi.org/10.18653/v1/2024.findings-eacl.88} {Goodhart{'}s law applies to {NLP}{'}s explanation benchmarks}.
\newblock In \emph{Findings of the Association for Computational Linguistics: EACL 2024}, pages 1322--1335, St. Julian{'}s, Malta. Association for Computational Linguistics.

\bibitem[{Hu et~al.(2025)Hu, Chen, and Ho}]{hu2025token}
Xiaomeng Hu, Pin-Yu Chen, and Tsung-Yi Ho. 2025.
\newblock Token highlighter: Inspecting and mitigating jailbreak prompts for large language models.
\newblock In \emph{Proceedings of the AAAI Conference on Artificial Intelligence}, volume~39, pages 27330--27338.

\bibitem[{Huang et~al.(2025)Huang, Huang, Yin, He, Zhen, Yuan, and Shao}]{huang2025attention}
Zixuan Huang, Kecheng Huang, Lihao Yin, Bowei He, Huiling Zhen, Mingxuan Yuan, and Zili Shao. 2025.
\newblock Attention-aware gnn-based input defense against multi-turn llm jailbreak.
\newblock \emph{arXiv preprint arXiv:2507.07146}.

\bibitem[{Husieva and Alvarez(2026)}]{alvarez2026geometry}
Yelyzaveta Husieva and Lauren Alvarez. 2026.
\newblock \href {https://arxiv.org/abs/2609.38389} {The geometry of harmfulness in multi-turn attacks}.
\newblock \emph{Preprint}, arXiv:2609.38389.

\bibitem[{Inan et~al.(2023)Inan, Upasani, Chi, Rungta, Iyer, Mao, Tontchev, Hu, Fuller, Testuggine et~al.}]{inan2023llama}
Hakan Inan, Kartikeya Upasani, Jianfeng Chi, Rashi Rungta, Krithika Iyer, Yuning Mao, Michael Tontchev, Qing Hu, Brian Fuller, Davide Testuggine, and 1 others. 2023.
\newblock Llama guard: Llm-based input-output safeguard for human-ai conversations.
\newblock \emph{arXiv preprint arXiv:2312.06674}.

\bibitem[{Jain and Wallace(2019)}]{jain-wallace-2019-attention}
Sarthak Jain and Byron~C. Wallace. 2019.
\newblock \href {https://doi.org/10.18653/v1/N19-1357} {{A}ttention is not {E}xplanation}.
\newblock In \emph{Proceedings of the 2019 Conference of the North {A}merican Chapter of the Association for Computational Linguistics: Human Language Technologies, Volume 1 (Long and Short Papers)}, pages 3543--3556, Minneapolis, Minnesota. Association for Computational Linguistics.

\bibitem[{Jiang et~al.(2023)Jiang, Sablayrolles, Mensch, Bamford, Chaplot, de~las Casas, Bressand, Lengyel, Lample, Saulnier, Lavaud, Lachaux, Stock, Le~Scao, Lavril, Wang, Lacroix, and El~Sayed}]{jiang2023mistral}
Albert~Q. Jiang, Alexandre Sablayrolles, Arthur Mensch, Chris Bamford, Devendra~Singh Chaplot, Diego de~las Casas, Florian Bressand, Gianna Lengyel, Guillaume Lample, Lucile Saulnier, L{\'e}lio~Renard Lavaud, Marie-Anne Lachaux, Pierre Stock, Teven Le~Scao, Thibaut Lavril, Thomas Wang, Timoth{\'e}e Lacroix, and William El~Sayed. 2023.
\newblock Mistral 7b.
\newblock \emph{arXiv preprint arXiv:2310.06825}.

\bibitem[{Jiang et~al.(2024)Jiang, Rao, Han, Ettinger, Brahman, Kumar, Mireshghallah, Lu, Sap, Choi et~al.}]{jiang2024wildteaming}
Liwei Jiang, Kavel Rao, Seungju Han, Allyson Ettinger, Faeze Brahman, Sachin Kumar, Niloofar Mireshghallah, Ximing Lu, Maarten Sap, Yejin Choi, and 1 others. 2024.
\newblock Wildteaming at scale: From in-the-wild jailbreaks to (adversarially) safer language models.
\newblock \emph{Advances in Neural Information Processing Systems}, 37:47094--47165.

\bibitem[{Kumar and Ahuja(2026)}]{kumar2026minimal}
Shubham Kumar and Narendra Ahuja. 2026.
\newblock Minimal, local, causal explanations for jailbreak success in large language models.
\newblock \emph{arXiv preprint arXiv:2605.00123}.

\bibitem[{Li et~al.(2024)Li, Han, Steneker, Primack, Goodside, Zhang, Wang, Menghini, and Yue}]{li2024llm}
Nathaniel Li, Ziwen Han, Ian Steneker, Willow Primack, Riley Goodside, Hugh Zhang, Zifan Wang, Cristina Menghini, and Summer Yue. 2024.
\newblock Llm defenses are not robust to multi-turn human jailbreaks yet.
\newblock \emph{arXiv preprint arXiv:2408.15221}.

\bibitem[{Lin et~al.(2023)Lin, Wang, Tong, Wang, Guo, Wang, and Shang}]{lin-etal-2023-toxicchat}
Zi~Lin, Zihan Wang, Yongqi Tong, Yangkun Wang, Yuxin Guo, Yujia Wang, and Jingbo Shang. 2023.
\newblock \href {https://doi.org/10.18653/v1/2023.findings-emnlp.311} {{T}oxic{C}hat: Unveiling hidden challenges of toxicity detection in real-world user-{AI} conversation}.
\newblock In \emph{Findings of the Association for Computational Linguistics: EMNLP 2023}, pages 4694--4702, Singapore. Association for Computational Linguistics.

\bibitem[{Liu et~al.(2026{\natexlab{a}})Liu, Hou, and Sui}]{liu-etal-2026-ramp}
Yize Liu, Yunyun Hou, and Aina Sui. 2026{\natexlab{a}}.
\newblock \href {https://doi.org/10.18653/v1/2026.findings-acl.925} {{RAMP}: Risk-aware multi-turn planning for jailbreak red-teaming}.
\newblock In \emph{Findings of the {A}ssociation for {C}omputational {L}inguistics: {ACL} 2026}, pages 18553--18564, San Diego, California, United States. Association for Computational Linguistics.

\bibitem[{Liu et~al.(2026{\natexlab{b}})Liu, Kim, Wan, Yuan, Tan, Mo, and Jiang}]{liu-etal-2026-mtmcs}
Zheyuan Liu, Dongwhi Kim, Yixin Wan, Xiangchi Yuan, Zhaoxuan Tan, Fengran Mo, and Meng Jiang. 2026{\natexlab{b}}.
\newblock \href {https://doi.org/10.18653/v1/2026.findings-acl.96} {{MTMCS}-bench: Evaluating contextual safety of multimodal large language models in multi-turn dialogues}.
\newblock In \emph{Findings of the {A}ssociation for {C}omputational {L}inguistics: {ACL} 2026}, pages 1992--2036, San Diego, California, United States. Association for Computational Linguistics.

\bibitem[{Lundberg and Lee(2017)}]{lundberg2017unified}
Scott~M Lundberg and Su-In Lee. 2017.
\newblock A unified approach to interpreting model predictions.
\newblock \emph{Advances in neural information processing systems}, 30.

\bibitem[{Mazeika et~al.(2024)Mazeika, Phan, Yin, Zou, Wang, Mu, Sakhaee, Li, Basart, Li et~al.}]{mazeika2024harmbench}
Mantas Mazeika, Long Phan, Xuwang Yin, Andy Zou, Zifan Wang, Norman Mu, Elham Sakhaee, Nathaniel Li, Steven Basart, Bo~Li, and 1 others. 2024.
\newblock Harmbench: A standardized evaluation framework for automated red teaming and robust refusal.
\newblock \emph{arXiv preprint arXiv:2402.04249}.

\bibitem[{Mehrotra et~al.(2024)Mehrotra, Zampetakis, Kassianik, Nelson, Anderson, Singer, and Karbasi}]{mehrotra2024tree}
Anay Mehrotra, Manolis Zampetakis, Paul Kassianik, Blaine Nelson, Hyrum Anderson, Yaron Singer, and Amin Karbasi. 2024.
\newblock Tree of attacks: Jailbreaking black-box llms automatically.
\newblock \emph{Advances in Neural Information Processing Systems}, 37:61065--61105.

\bibitem[{{Mistral AI}(2025)}]{mistral2025small31}
{Mistral AI}. 2025.
\newblock Mistral small 3.1.
\newblock \url{https://mistral.ai/news/mistral-small-3-1/}.
\newblock Mistral-Small-3.1-24B-Instruct-2503.

\bibitem[{{OpenAI}(2026{\natexlab{a}})}]{openai2026gpt56}
{OpenAI}. 2026{\natexlab{a}}.
\newblock {GPT-5.6}: Frontier intelligence that scales with your ambition.
\newblock \url{https://openai.com/index/gpt-5-6/}.

\bibitem[{{OpenAI}(2026{\natexlab{b}})}]{openai2026gpt6astra}
{OpenAI}. 2026{\natexlab{b}}.
\newblock {GPT-6 Astra}: A new generation of intelligence.
\newblock \url{https://openai.com/index/gpt-6-astra/}.

\bibitem[{{Qwen Team}(2025)}]{qwen2025qwen25}
{Qwen Team}. 2025.
\newblock \href {https://doi.org/10.48550/arXiv.2412.15115} {Qwen2.5 technical report}.
\newblock \emph{arXiv preprint arXiv:2412.15115}.

\bibitem[{R{\"o}ttger et~al.(2024)R{\"o}ttger, Kirk, Vidgen, Attanasio, Bianchi, and Hovy}]{rottger-etal-2024-xstest}
Paul R{\"o}ttger, Hannah Kirk, Bertie Vidgen, Giuseppe Attanasio, Federico Bianchi, and Dirk Hovy. 2024.
\newblock \href {https://doi.org/10.18653/v1/2024.naacl-long.301} {{XST}est: A test suite for identifying exaggerated safety behaviours in large language models}.
\newblock In \emph{Proceedings of the 2024 Conference of the North American Chapter of the Association for Computational Linguistics: Human Language Technologies (Volume 1: Long Papers)}, pages 5377--5400, Mexico City, Mexico. Association for Computational Linguistics.

\bibitem[{Russinovich et~al.(2025)Russinovich, Salem, and Eldan}]{russinovich2025great}
Mark Russinovich, Ahmed Salem, and Ronen Eldan. 2025.
\newblock Great, now write an article about that: The {Crescendo} multi-turn {LLM} jailbreak attack.
\newblock In \emph{34th USENIX Security Symposium (USENIX Security 25)}, pages 2421--2440.

\bibitem[{Song et~al.(2026)Song, Liu, Yang, Chen, Feng, Zhu, and Gao}]{song2026multibreak}
Jialin Song, Xiaodong Liu, Weiwei Yang, Wuyang Chen, Mingqian Feng, Xuekai Zhu, and Jianfeng Gao. 2026.
\newblock Multibreak: A scalable and diverse multi-turn jailbreak benchmark for evaluating llm safety.
\newblock \emph{arXiv preprint arXiv:2605.01687}.

\bibitem[{Sundararajan et~al.(2017)Sundararajan, Taly, and Yan}]{sundararajan2017axiomatic}
Mukund Sundararajan, Ankur Taly, and Qiqi Yan. 2017.
\newblock Axiomatic attribution for deep networks.
\newblock In \emph{International conference on machine learning}, pages 3319--3328. PMLR.

\bibitem[{Warner et~al.(2024)Warner, Chaffin, Clavié, Weller, Hallström, Taghadouini, Gallagher, Biswas, Ladhak, Aarsen, Cooper, Adams, Howard, and Poli}]{warner2024smarterbetterfasterlonger}
Benjamin Warner, Antoine Chaffin, Benjamin Clavié, Orion Weller, Oskar Hallström, Said Taghadouini, Alexis Gallagher, Raja Biswas, Faisal Ladhak, Tom Aarsen, Nathan Cooper, Griffin Adams, Jeremy Howard, and Iacopo Poli. 2024.
\newblock \href {https://arxiv.org/abs/2412.13663} {Smarter, better, faster, longer: A modern bidirectional encoder for fast, memory efficient, and long context finetuning and inference}.
\newblock \emph{Preprint}, arXiv:2412.13663.

\bibitem[{Xiong et~al.(2026)Xiong, Li, Liu, Li, Shi, Zhu, Wang, and Feng}]{xiong-etal-2026-trojail}
Xiqiao Xiong, Ouxiang Li, Zhuo Liu, Moxin Li, Wentao Shi, Fengbin Zhu, Qifan Wang, and Fuli Feng. 2026.
\newblock \href {https://doi.org/10.18653/v1/2026.acl-long.2220} {{TROJ}ail: Trajectory-level optimization for multi-turn large language model jailbreaks with process rewards}.
\newblock In \emph{Proceedings of the 64th Annual Meeting of the {A}ssociation for {C}omputational {L}inguistics (Volume 1: Long Papers)}, pages 48086--48109, San Diego, California, United States. Association for Computational Linguistics.

\bibitem[{Yu et~al.(2024)Yu, Li, Liao, Wang, Gao, Mi, and Hong}]{yu-etal-2024-cosafe}
Erxin Yu, Jing Li, Ming Liao, Siqi Wang, Zuchen Gao, Fei Mi, and Lanqing Hong. 2024.
\newblock \href {https://doi.org/10.18653/v1/2024.emnlp-main.968} {{C}o{S}afe: Evaluating large language model safety in multi-turn dialogue coreference}.
\newblock In \emph{Proceedings of the 2024 Conference on Empirical Methods in Natural Language Processing}, pages 17494--17508, Miami, Florida, USA. Association for Computational Linguistics.

\bibitem[{Zeng et~al.(2024)Zeng, Liu, Mullins, Peran, Fernandez, Harkous, Narasimhan, Proud, Kumar, Radharapu et~al.}]{zeng2024shieldgemma}
Wenjun Zeng, Yuchi Liu, Ryan Mullins, Ludovic Peran, Joe Fernandez, Hamza Harkous, Karthik Narasimhan, Drew Proud, Piyush Kumar, Bhaktipriya Radharapu, and 1 others. 2024.
\newblock Shieldgemma: Generative ai content moderation based on gemma.
\newblock \emph{arXiv preprint arXiv:2407.21772}.

\bibitem[{Zhang et~al.(2025)Zhang, Zhai, Guo, Hu, Guo, Fang, Zhao, Shen, Wang, and Wang}]{zhang2025jbshield}
Shenyi Zhang, Yuchen Zhai, Keyan Guo, Hongxin Hu, Shengnan Guo, Zheng Fang, Lingchen Zhao, Chao Shen, Cong Wang, and Qian Wang. 2025.
\newblock {JBShield}: Defending large language models from jailbreak attacks through activated concept analysis and manipulation.
\newblock In \emph{34th USENIX Security Symposium (USENIX Security 25)}, pages 8215--8234.

\bibitem[{Zou et~al.(2023)Zou, Wang, Carlini, Nasr, Kolter, and Fredrikson}]{zou2023universal}
Andy Zou, Zifan Wang, Nicholas Carlini, Milad Nasr, J~Zico Kolter, and Matt Fredrikson. 2023.
\newblock Universal and transferable adversarial attacks on aligned language models.
\newblock \emph{arXiv preprint arXiv:2307.15043}.

\end{thebibliography}
\appendix
\section{Dataset Generation Pipeline}
\label{app:dataset_generation}
\subsection{Generation and Validation Configuration}
\label{app:generation-config}

\begin{figure}[!htbp]
    \centering
    \includegraphics[width=\linewidth]{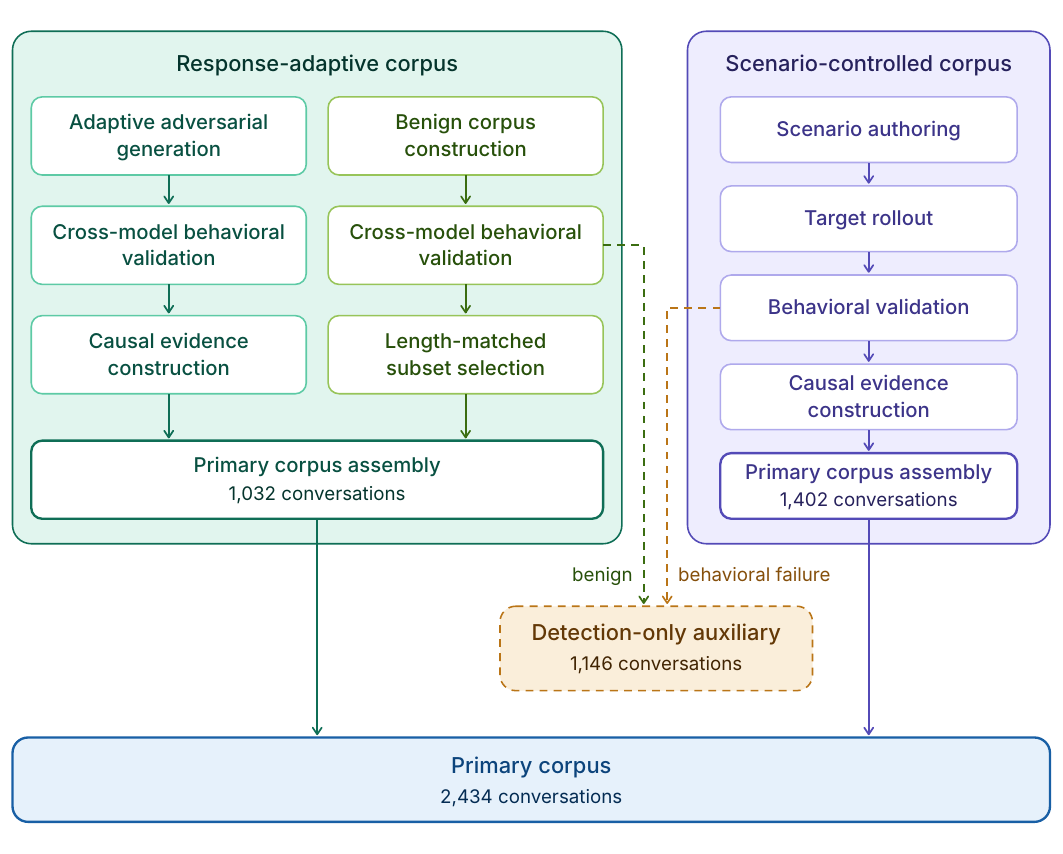}
    \caption{
        Construction of the primary corpus and detection-only auxiliary data. Solid paths contribute to the primary corpus, and dashed paths contribute detection-only auxiliary examples.
    }
    \label{fig:corpus_pipeline}
\end{figure}

Figure~\ref{fig:corpus_pipeline} summarizes the construction of \textsc{MTAG-Bench}. Table~\ref{tab:generation-config} summarizes the models utilized. Response-adaptive trajectories use Qwen2.5 generators~\citep{qwen2025qwen25}, Llama-3-8B-Instruct as target~\citep{grattafiori2024llama}, and Mistral-7B-Instruct-v0.3 for validation and counterfactual judging~\citep{jiang2023mistral}. Scenario-controlled trajectories use GPT-5.6 Sol/GPT-6 Astra for authoring~\citep{openai2026gpt56,openai2026gpt6astra}, Qwen2.5-32B-Instruct as the target~\citep{qwen2025qwen25}, and Mistral-Small-3.1-24B-Instruct as the validator and counterfactual judge~\citep{mistral2025small31}. Counterfactual evidence replay uses the same target with deterministic decoding.

\begin{table}[!htbp]
\centering
\small
\setlength{\tabcolsep}{1pt}
\renewcommand{\arraystretch}{1.08}
\begin{tabular}{p{0.38\columnwidth} p{0.58\columnwidth}}
\toprule
\textbf{Role} & \textbf{Model} \\
\midrule

\multicolumn{2}{l}{\textit{Response-adaptive}} \\
Adversarial generator
& Qwen2.5-7B/14B-Instruct \\
Benign generator
& Qwen2.5-7B-Instruct \\
Target / CF replay
& Llama-3-8B-Instruct \\
Validator / CF judge
& Mistral-7B-Instruct-v0.3 \\
\midrule

\multicolumn{2}{l}{\textit{Scenario-controlled}} \\
Trajectory generators
& OpenAI GPT-5.6 and GPT-6 \\
Target / CF replay
& Qwen2.5-32B-Instruct \\
Validator / CF judge
& Mistral-Small-3.1-24B-Instruct \\
\midrule
\multicolumn{2}{l}{\textit{Behavioral replay}} \\
Source-specific judge & Corpus validator (as above) \\
Common judge & Gemma-4-31B-IT \\
\bottomrule
\end{tabular}
\caption{Models used for dataset construction, validation and behavioral replay.}
\label{tab:generation-config}
\end{table}

Scenario-controlled rollout, judging, and counterfactual replay use vLLM~0.28.0 on A100 80\,GB GPUs in BF16. Execution is eager and batch-invariant. The target uses a 16,384-token context for the factual rollout. Counterfactual replay can use the 32,768-token context envelope required by the adaptive completion policy. The safety judge uses a 32,768-token context and receives the complete observable context.

For scenario-controlled trajectories, a response is considered confidently unsafe when its unsafe score is at least $0.50$ and the judge's confidence is at least $0.55$. Scores in $[0.35,0.50)$, as well as high-risk judgements below the confidence threshold, are treated as ambiguous. The earliest confidently unsafe response defines the intervention anchor. Candidate materialization considers at most four user turns. Counterfactual replay evaluates at most six candidate evidence spans and two preceding-context negative controls per record.

\paragraph{Pipeline label validation.}
\label{app:label_validation}
All 182 unsafe test conversations carry a pipeline unsafe label from one of three supervision tiers. Table~\ref{tab:label_validation} reports how often both annotators judge conversations unsafe or partially unsafe. Counterfactual-strong records are confirmed at roughly twice the rate of LLM-confirmed records (84.2\% vs.\ 41.9\%; Fisher exact $p \approx 2.4\times10^{-9}$). The LLM-confirmed tier therefore carries substantially noisier trajectory-level labels. These records contribute to conversation-level detection, but do not receive positive token labels. They receive a positive turn label only when a turn-level intervention on that record is independently supported. Behavioral replay and human validation use cases with reproduced or human-confirmed unsafe behavior, reducing their dependence on pipeline labels. Detection and internal intervention metrics use the entire unsafe test cohort. Of the 117 test conversations with positive turn evidence, 96 are counterfactually supported records and 21 are LLM-confirmed records with an independently supported turn intervention. Among the 21 LLM-confirmed records, both annotators judged the realized trajectory unsafe or partially unsafe in 12 cases, one annotator did so in 2 cases, and neither did in 7 cases. Span-annotated conversation counts include any record with at least one span annotation, including benign controls, rather than only records with counterfactually supported evidence.

\subsection{Neutralization techniques}
\label{app:neutralization}
Counterfactual replacements are sampled with a seeded RNG from fixed neutralization pools. Whole-turn replacements are \{``Can you summarize the main points we have discussed so far?'', ``Can we step back and review the key ideas from this conversation?'', ``Please give a concise overview of the discussion up to this point.'', ``What are the main considerations we have covered so far?''\}. Evidence spans use \{``the relevant part of the request'', ``the general topic being discussed'', ``that part of the earlier context''\}, while negative-control spans use \{``in this context'', ``for the current discussion'', ``as background context''\}. Response-adaptive counterfactuals use analogous fixed, span-type-specific neutralization pools.

\begin{table}[t]
\centering
\small
\setlength{\tabcolsep}{1pt}
\renewcommand{\arraystretch}{1.08}
\begin{tabular}{lcccc}
\toprule
\textbf{Statistic} & \textbf{Train} & \textbf{Dev} & \textbf{Test} & \textbf{Total} \\
\midrule

\multicolumn{5}{l}{\textit{Primary benchmark}} \\
Avg.\ primary turns    & 17.60 & 17.36 & 17.50 & 17.55 \\
Avg.\ primary user turns &  8.80 &  8.68 &  8.75 &  8.77 \\
\midrule

\multicolumn{5}{l}{\textit{Corpus sources}} \\
Scenario-controlled                     & 980 & 212 & 210 & 1,402 \\
Response-adaptive                       & 726 & 152 & 154 & 1,032 \\
\midrule

\multicolumn{5}{l}{\textit{Conversation families}} \\
Scenario-controlled adversarial         & 490 & 106 & 105 & 701 \\
Scenario-controlled benign              & 490 & 106 & 105 & 701 \\
Response-adaptive adversarial               & 363 &  76 &  77 & 516 \\
Response-adaptive benign twin               & 363 &  76 &  77 & 516 \\
\midrule

\multicolumn{5}{l}{\textit{Evidence supervision}} \\
Turn-evidence labeled                 & 582 & 125 & 117 & 824 \\
Turn-evidence annotations             & 1,081 & 213 & 228 & 1,522 \\
Span-annotated conversations                & 895 & 194 & 196 & 1,285 \\
Total span annotations                & 4,077 & 900 & 873 & 5,850 \\
CF-supported spans              & 844 & 146 & 182 & 1,172 \\
Incidental spans              & 221 &  48 &  46 & 315 \\
Unvalidated/ignored spans             & 3,012 & 706 & 645 & 4,363 \\
\midrule

\multicolumn{5}{l}{\textit{Span annotation types}} \\
Evidence candidate                    & 3,800 & 826 & 804 & 5,430 \\
Control candidate                     &   212 &  42 &  46 &   300 \\
Safe constraint                       &    61 &  32 &  20 &   113 \\
Payload span                          &     3 &   0 &   2 &     5 \\
Structural trigger                    &     1 &   0 &   1 &     2 \\
\midrule

\multicolumn{5}{l}{\textit{Detection-only training auxiliaries}} \\
Response-adaptive auxiliary                      & 721 & -- & -- & 721 \\
Scenario-controlled auxiliary                    & 425 & -- & -- & 425 \\
\bottomrule
\end{tabular}
\caption{Additional MTAG-Bench dataset statistics.}
\label{tab:dataset_detailed}
\end{table}

\subsection{Failure Handling and Exclusions}
\label{app:failure-handling}

Independent-validation failures, such as negative outcomes and judge or runtime errors, are not included for supervision. Scenario-controlled judge responses receive up to three structured parsing attempts per rubric pass. Any baseline mismatch in counterfactual validation makes the corresponding intervention invalid. Counterfactuals exceeding the permitted context envelope are excluded from localization supervision. For scenario-controlled generation, target completions that terminate at the 2,048-token output limit are reevaluated at 4,096 and 8,192 tokens. Response-adaptive adversarial attempts that do not produce a validated jailbreak are retained only as stress negatives and are not treated as benign. 

\subsection{Dataset Validity Checks}
\label{app:dataset-validity}

\paragraph{Human attribution validation.}
We perform human validation on the held-out test set using two independent annotators. The study uses the exact top-15\% user-word attribution produced by MTAG-Probe. Annotators are blind to conversation identifiers, supervision and counterfactual status, judge outputs, model probabilities, corpus source, gold evidence annotations, and model identity. Both annotators independently assess all 182 conversations in different deterministic orders, judging the realized assistant outcome, attribution support, evidence precision and coverage, essential or incorrect attributed evidence, and important missed user evidence. Inter-annotator agreement and per-annotator statistics are computed before any adjudication.

\begin{table}[!htbp]
\centering
\small
\setlength{\tabcolsep}{4pt}
\begin{tabular}{@{}lrccrr@{}}
\toprule
\textbf{Tier} & \textbf{$N$} & \textbf{Both} & \textbf{Both (\%)} & \textbf{One} & \textbf{Neither} \\
\midrule
CF strong     & 95  & 80  & 84.2 & 7  & 8  \\
LLM-confirmed & 86  & 36  & 41.9 & 10 & 40 \\
CF weak       & 1   & 1   & --   & 0  & 0  \\
\midrule
Overall       & 182 & 117 & 64.3 & 17 & 48 \\
\bottomrule
\end{tabular}
\caption{Human validation of pipeline unsafe labels.}
\label{tab:label_validation}
\footnotesize \raggedright
Each unsafe test conversation is counted by how many of the two annotators independently judged its outcome unsafe or partially unsafe. Wilson 95\% CIs: CF strong [75.6, 90.2], LLM-confirmed [32.0, 52.4], overall [57.1, 70.9]. CF strong vs.\ LLM-confirmed: Fisher exact $p \approx 2.4\times10^{-9}$.
\end{table}

\subsection{Splits and auxiliary data}
\label{app:data-splits}
We use a group-aware 70/15/15 split. The paired conversations and related scenario families remain within the same partition to prevent the shared context from crossing the train, development, and test boundaries. Records with identical normalized user-turn trajectories are assigned to the same split and source corpus, regardless of differences in record identifiers or scenario-family labels. A pair enters the primary corpus only when both its adversarial and benign trajectories independently pass their respective validation gate. Detection-only auxiliary examples are restricted to training.

\section{MTAG-Probe Implementation Details}
\label{app:model-details}

\subsection{Architecture Configuration}
\label{app:model-architecture}

MTAG-Probe uses ModernBERT as its turn encoder, with 1,024-dimensional token representations and the backbone revision fixed for all reported experiments. Each realized conversation turn is encoded independently at its native length, with a maximum supported length of 8,192 tokens per turn and a maximum of 64 turns per conversation. Turns that exceed the supported representation limits are not considered for training or evaluation.

Within each turn, learned attention pooling maps the ModernBERT token representations to a single turn vector. The pooled representation is projected from 1,024 to 256 dimensions and augmented with a fixed sinusoidal turn-position encoding and a learned speaker-role embedding. Cross-turn contextualization uses a two-layer Transformer with eight attention heads, a $4\times$ feed-forward expansion, GELU activations, pre-normalization, and dropout $0.1$. A final layer normalization produces the contextualized turn representations. Since cross-turn attention operates over one representation per turn, its attention cost scales as $O(T^2)$ in the number of turns.

A learned attention pool over the contextualized turn representations produces the conversation representation used by the detection head. The detection head is an MLP with two hidden layers of width 256 and 128, GELU activations, and dropout $0.1$. The turn localization uses an independent two-layer MLP ($256\!\rightarrow\!128\!\rightarrow\!1$) applied to each contextualized turn representation.

For token localization, the original 1,024-dimensional ModernBERT token representation is projected to 128 dimensions, while the contextualized 256-dimensional representation of its containing turn is independently projected to 128 dimensions. The two representations are summed and passed through layer normalization, GELU, dropout, and a binary prediction layer. Thus, turn scores are not obtained by aggregating token scores, and token attribution is conditioned on the conversation-aware state of the containing turn. The assistant and padded positions are masked from both turn- and token-localization outputs.

\paragraph{Representation Coverage.}
The maximum realized turn lengths are 3,491 tokens in training and 2,113 tokens in development, both below ModernBERT's 8,192-token context limit. No train or development turn exceeded the configured token ceiling.

\paragraph{Parameter count.}
MTAG-Probe adds 2.30M task-specific parameters on top of the ModernBERT-large encoder. The complete model contains 397.09M parameters, of which 51.33M (12.93\%) are trainable. The trainable set consists of 49.03M parameters from the final four ModernBERT layers and final normalization layer, and 2.30M task-specific parameters.

\begin{table}[!htbp]
\centering
\small
\setlength{\tabcolsep}{4pt}
\begin{tabular}{lr}
\toprule
\textbf{MTAG-Probe component} & \textbf{Parameters} \\
\midrule
Cross-turn encoder & 1,974,273 \\
Token-localization head & 164,481 \\
Detection head & 98,817 \\
Conversation pooler & 33,025 \\
Turn-localization head & 33,025 \\
\midrule
\textbf{Total MTAG-Probe} & \textbf{2,303,621} \\
\bottomrule
\end{tabular}
\caption{Task-specific parameter count for MTAG-Probe.}
\label{tab:mtag_parameters}
\end{table}

\subsection{Supervision and Loss Weighting}
\label{app:model-loss}

All three objectives use binary cross-entropy. Strong counterfactual evidence receives localization weight $1.0$, while weak counterfactual evidence receives weight $0.70$. Validated benign examples and explicitly supported negative controls receive unit negative weight. 

Confidence weights are applied as absolute loss multipliers and are not renormalized within each batch. For conversation-level detection, the positive-class weight is computed from the weighted negative-to-positive training mass. The turn-localization loss uses the analogous ratio over supervised turn targets. Token localization does not use any additional positive-class reweighting.

Detection-only auxiliary examples do not contribute turn or token supervision. Response-adaptive auxiliary examples receive detection weight $1.0$, while scenario-controlled auxiliary examples receive detection weight $0.25$. Behaviorally validated unsafe primary examples without supported local evidence contribute to conversation-level detection only and are excluded from localization supervision.

\subsection{Optimization}
\label{app:model-optimization}

The final model fine-tunes the last four Transformer layers and the final normalization layer of ModernBERT-large while keeping the remaining backbone parameters frozen. The four trainable Transformer layers, the final normalization layer, and the task-specific parameters remain in FP32, while CUDA autocast uses BF16 activations. The frozen backbone prefix is stored in BF16, and gradient checkpointing is enabled for the selectively trainable backbone layers.

We train for 20 epochs using AdamW. Task-specific parameters use a learning rate of $2\times10^{-4}$ and trainable ModernBERT parameters use $2\times10^{-5}$, with weight decay $0.01$. A one-cycle learning-rate schedule with cosine annealing is used, with the increasing phase spanning 200 optimizer steps. Training uses batch size 2, eight-step gradient accumulation, gradient clipping at $1.0$, and seed 42.

\subsection{Architecture Selection}
\label{sec:architecture-selection}

We compared six retrospective attribution architectures on the development set. The candidates varied in cross-turn aggregation, interaction mechanism, and whether the ModernBERT backbone was frozen or its top four layers were 
fine-tuned. Architecture selection considered detection, localization, and 
development-set intervention behavior. It was separate from the per-architecture checkpoint rule described below.

\begin{table*}[t]
\centering
\small
\setlength{\tabcolsep}{3.5pt}
\renewcommand{\arraystretch}{1.08}
\begin{tabular}{@{}lrrrrrcc@{}}
\toprule
\textbf{Model} &
\textbf{Det.\ AP} &
\textbf{Turn AP} &
\textbf{Span F1} &
$\boldsymbol{\Delta P@15}$ &
\textbf{Flip@15} &
\shortstack{\textbf{$\Delta$Flip vs.}\\\textbf{LOTO (pp)}} &
\shortstack{\textbf{$\Delta$Flip vs.}\\\textbf{SpanRand (pp)}} \\
\midrule
Cross-token-gated-frozen
& 94.3 & 92.0 & 95.7 & 0.1 & 3.3 & $-2.2$ & $-0.3$ \\
Cross-token-gated-top4
& 97.2 & 90.7 & 95.4 & 15.9 & 29.7 & $+1.0$ & $+1.7$ \\
Cross-token-sibling-frozen
& 96.4 & 93.3 & 97.0 & 19.6 & 24.7 & $+3.1$ & $+4.2$ \\
Cross-token-sibling-top4
& 97.5 & 94.6 & 97.0 & 4.0 & 9.9 & $-5.1$ & $-0.6$ \\
Hierarchical-sibling-frozen
& 95.0 & 92.1 & 96.7 & 30.5 & 45.6 & $+1.6$ & $+0.4$ \\
\textbf{Hierarchical-sibling-top4}
& 96.1
& 90.6
& 94.9
& \textbf{36.6}
& 41.2
& $\mathbf{+7.8}$
& $\mathbf{+5.0}$ \\
\bottomrule
\end{tabular}
\caption{Development-set architecture comparison.}
\label{tab:architecture-selection}
\footnotesize 
AP, F1, and Flip values are percentages. $\Delta P@15$ is the absolute decrease in unsafe probability, expressed in percentage points, at a 15\% user-word attribution budget. Flip contrasts are percentage-point differences in flip rate. SpanRand denotes randomized-span deletion. Each candidate uses its model-specific
development threshold.
\end{table*}

We selected the hierarchical sibling model with the top four backbone layers 
fine-tuned, which we denote MTAG-Probe. It achieved the largest 
$\Delta P@15$ among the six candidates. Its Flip@15 exceeded LOTO by 7.8 
percentage points and randomized-span deletion by 5.0 points. Against matched 
random deletion, the corresponding improvement was 29.0 points.

The closest trainable alternative was the cross-token sibling model. In a 
paired comparison, MTAG-Probe produced significantly more flips at 
every attribution budget.

\begin{table}[t]
\centering
\small
\begin{tabular}{rcc}
\toprule
Budget & $\Delta$Flip & 95\% CI \\
\midrule
5\%  & $+15.4$ pp & $[9.3,\ 22.0]$ \\
10\% & $+20.9$ pp & $[13.2,\ 28.6]$ \\
15\% & $+31.3$ pp & $[23.6,\ 39.0]$ \\
20\% & $+35.2$ pp & $[26.9,\ 43.4]$ \\
\bottomrule
\end{tabular}
\caption{
Paired development-set flip-rate difference between the hierarchical sibling and cross-token 
sibling top-four models.}
\label{tab:hierarchical-cross-token}
\end{table}

The top-four variant did not dominate its frozen counterpart on every metric. 
The frozen model had higher raw flip rates at the 10\%, 15\%, and 20\% budgets. However, it was significantly worse than LOTO at the 10\% budget 
($-4.3$ points, 95\% CI $[-7.4,-1.3]$). The top-four model produced larger 
probability drops at all four budgets and stronger baseline-adjusted flip 
effects. It also increased detection AP from 95.0\% to 96.1\%, reduced the 
development false-positive rate from 24.2\% to 18.1\%, and increased turn F1 
from 55.9\% to 58.8\%. We therefore selected it as the better overall 
detector-localizer configuration.

\subsection{Checkpoint Selection and Calibration}
\label{app:model-selection}

All checkpoint selection and operating-point calibration use only the development set. The MTAG-Probe checkpoint is selected by the mean of development-set turn and token localization average precision, ensuring that checkpoint selection reflects both levels of evidence localization.

For conversation-level detection, the final operating point is calibrated separately on the scenario-controlled development subset. We choose the threshold with maximum recall subject to a false-positive rate of at most $5\%$. The resulting threshold is frozen before evaluation on the held-out test set. Turn- and token-level binary localization metrics use a threshold of $0.5$, while ranking, top-$k$, and intervention evaluations operate directly on continuous attribution scores.

For each architecture, we trained for 20 epochs, evaluated after every epoch, and 
retained the epoch that maximized the mean of development turn AUPRC and span 
AUPRC:
\begin{equation}
e^{*}
=
\arg\max_{e}
\frac{
    \mathrm{AUPRC}^{(e)}_{\mathrm{turn}}
    +
    \mathrm{AUPRC}^{(e)}_{\mathrm{span}}
}{2}.
\end{equation}
Detection F1 and intervention outcomes did not determine the retained epoch. 
For the selected architecture, this rule selected epoch~8. The held-out test set remains isolated from training, checkpoint selection, and threshold calibration.

The flip results in Table~\ref{tab:architecture-selection} were computed at 
candidate-specific development thresholds. The selected model used a threshold 
of 0.445 during architecture selection. Final held-out evaluation used the 
separately fixed threshold of 0.6451. Because flip rate depends on the decision threshold, development-selection flip rates and final held-out flip rates are reported as distinct results.

\subsection{Intervention Construction and Lexical Control}
\label{app:intervention_details}
\paragraph{Intervention unit and deletion operator.}
MTAG-Probe produces attribution scores over model tokens, whereas intervention budgets are defined over user-visible \emph{word units}. We define a word unit as a maximal non-whitespace span in a user message. Thus, punctuation attached to a word remains part of the same intervention unit. For each word $w_i$, we project the model-token attribution scores to the word level using the maximum score among all model tokens whose character offsets overlap $w_i$:
\[
s(w_i)=\max_{j:\,\mathrm{span}(t_j)\cap\mathrm{span}(w_i)\neq\emptyset}
s(t_j).
\]
Only words in user turns are eligible for the primary intervention.

For a budget $b$ and $N$ eligible user words, each method selects
\[
k=\min(N,\lceil bN\rceil)
\]
words. The reported 5\%, 10\%, 15\%, and 20\% intervention budgets therefore
refer to fractions of eligible user words, rather than tokenizer subwords. All methods compared at a given budget remove exactly the same number of words.

The primary intervention is textual deletion. Selected word spans are removed from the original user text using their character offsets, after which the edited conversation is passed through the model's standard tokenizer again. We do not replace selected content with padding tokens, mask tokens, zero embeddings, or zeroed activations. In static intervention experiments, assistant turns are held fixed while the edited trajectory is rescored. In full-trajectory behavioral replay, we first delete the selected user words. Starting from the earliest user turn affected by any intervention, we then regenerate all subsequent assistant responses while keeping the remaining user turns unchanged.

We additionally evaluate a blanking sensitivity control in which each character
of a selected word is replaced with a space rather than deleted, preserving the
raw character length before normal re-tokenization. This operation does not use
padding tokens, mask tokens, or zeroed representations. At the 15\% user-word
budget, blanking produces a 50.0\% [42.3\%, 57.1\%] unsafe-to-safe flip rate, compared with 48.4\% under the primary deletion intervention. The similar intervention effect indicates that the result is not driven by the specific choice between deleting and blanking the selected text.

\paragraph{SurfaceRisk construction.}
SurfaceRisk is an expanded lexical-risk control used to test whether MTAG-Probe
primarily identifies overtly hazardous surface language. The frozen
specification contains 113 exact lexical terms, 27 morphological prefix stems,
and 41 multiword phrases. Its vocabulary was externally motivated by OWASP
LLM01:2025 Prompt Injection, MITRE ATT\&CK Enterprise tactics and techniques,
HarmBench semantic harm categories, and the ShieldGemma dangerous-content
categories already fixed for our evaluation.

Matching is case-normalized. The exact terms and morphological prefixes receive
graded risk scores, while listed multiword phrases receive the maximum lexical
risk score. Phrase matching is restricted to a single user turn. To capture
immediate lexical context, the word directly adjacent to either side of a
matched phrase also receives a lower risk score. When multiple rules apply to
a word, the maximum applicable score is used.

For intervention experiments, SurfaceRisk ranks the same eligible user-word
population used by MTAG-Probe and removes the highest-ranked $k$ words under
the same budget. Deterministic turn and character-offset ordering resolves
score ties, ensuring an exact matched deletion count even when fewer than $k$
words receive nonzero lexical scores.

\begin{table}[t]
\centering
\small
\setlength{\tabcolsep}{3.5pt}
\begin{tabular}{p{0.19\columnwidth} p{0.22\columnwidth} p{0.50\columnwidth}}
\toprule
\textbf{Method} & \textbf{Type} & \textbf{Purpose} \\
\midrule
LOTO
& Baseline
& Masks each user turn independently and ranks turns by the resulting decrease in unsafe probability from the same conversation classifier. \\

Matched random
& Baseline
& Deletes an equal number of user words at arbitrary locations, testing whether comparable effects arise from deletion alone. \\

SurfaceRisk
& Baseline
& Ranks user evidence using lexical surface-risk cues, testing whether localization can be recovered from risky wording alone. \\

LLM attribution
& Baseline
& Asks the failed target model and an independent safety judge to directly identify evidence-bearing user turns and spans. \\

Last-user
& Baseline
& Attributes evidence to the most recent user request, testing a simple recency heuristic. \\
\midrule

MTAG-span-random
& Control
& Preserves MTAG-Probe's selected turns, per-turn deletion allocation, and contiguous-span lengths while randomizing span positions within those turns. \\

Context-only
& Control
& Protects the final user request and restricts MTAG-Probe's intervention to preceding user context. \\
\bottomrule
\end{tabular}
\caption{Baselines and intervention controls.}
\label{tab:baselines-interventions}
\end{table}

\section{Additional Evaluation Results}
\label{app:full_results}
\subsection{Full-Trajectory Behavioral Replay}
\label{app:replay_results}

 Table~\ref{tab:replay_detailed} reports record-level behavioral replay results under the source-specific and common independent judges. The analysis conditions on conversations whose original replayed final response is judged unsafe, yielding 99 eligible records under the source-specific judge and 51 under the common judge. At the 15\% user-word deletion budget, MTAG-Probe changes 84.3\% of eligible final responses from unsafe to safe under both protocols, substantially exceeding matched random and SurfaceRisk deletion.

\begin{table}[!htbp]
\centering
\small
\setlength{\tabcolsep}{2.0pt}
\renewcommand{\arraystretch}{1.08}
\begin{tabular}{@{}lcc@{}}
\toprule
\textbf{Method} &
\textbf{Source-specific judge} &
\textbf{Common judge} \\
\midrule
MTAG-Probe &
\textbf{84.3 [76.8, 90.9]} &
\textbf{84.3 [74.5, 92.2]} \\

Random &
49.5 [39.4, 59.6] &
43.1 [29.4, 56.9] \\

SurfaceRisk &
33.5 [24.4, 42.4] &
39.9 [26.1, 52.9] \\
\bottomrule
\end{tabular}
\caption{Full-trajectory behavioral replay.}
\label{tab:replay_detailed}
\footnotesize
Values are final-response unsafe$\rightarrow$safe transition rates at a 15\% user-word deletion budget. Confidence intervals are cluster-bootstrap 95\% intervals.
\end{table}

Under the source-specific judge, MTAG-Probe exceeds random deletion by 34.8 points (95\% CI [24.2, 45.5]) and SurfaceRisk by 50.8 points ([39.6, 62.1]). Under the common judge, the corresponding differences are 41.2 points ([23.5, 56.9]) and 44.4 points ([27.5, 61.4]). All four intervals exclude zero. Under the stricter downstream-trajectory criterion, MTAG-Probe produces unsafe-to-safe transitions in 55.8\% and 70.4\% of eligible trajectories under the source-specific and common judges, respectively, compared with 27.4\% and 27.6\% for matched random deletion and 28.8\% and 40.3\% for SurfaceRisk.

\subsection{Fixed-Budget Intervention Faithfulness}
\label{app:internal_intervention}

Table~\ref{tab:internal_intervention_detailed} reports the exact unsafe$\rightarrow$safe transition rates underlying Figure~\ref{fig:internal_attribution}(a). Across budgets from 5\% to 20\% of user words, MTAG-Probe produces larger decision changes than LOTO, matched random deletion and SurfaceRisk. 

\begin{table}[!htbp]
\centering
\small
\setlength{\tabcolsep}{3.4pt}
\renewcommand{\arraystretch}{1.08}
\begin{tabular}{@{}lrrrr@{}}
\toprule
\textbf{Method} &
\textbf{5\%} &
\textbf{10\%} &
\textbf{15\%} &
\textbf{20\%} \\
\midrule
MTAG-Probe &
\textbf{26.9} &
\textbf{40.1} &
\textbf{48.4} &
\textbf{49.5} \\

LOTO &
15.4 & 32.4 & 36.8 & 38.5 \\

MTAG-span-random &
23.1 & 35.8 & 44.0 & 49.2 \\

Matched random &
3.6 & 8.4 & 11.2 & 15.5 \\

SurfaceRisk &
1.1 & 1.1 & 1.1 & 1.1 \\

Last-user &
1.1 & 2.7 & 2.2 & 2.7 \\
\bottomrule
\end{tabular}
\caption{Internal fixed-budget intervention results.}
\label{tab:internal_intervention_detailed}
\footnotesize \raggedright
Values are unsafe$\rightarrow$safe decision-transition rates on the
182 held-out unsafe conversations. Budgets denote the fraction of
user words removed.
\end{table}

MTAG-span-random preserves the MTAG-Probe-selected turns, deletion allocation, and span lengths while randomizing the within-turn span positions. Its relatively strong performance shows that identifying the relevant turns accounts for a substantial portion of the intervention effect, while MTAG-Probe's within-turn localization provides additional benefit at the primary 15\% budget. In contrast, restricting deletion to the final user request has little effect, indicating that the intervention results cannot be explained by recency alone.

\begin{table*}[t]
\centering
\small
\setlength{\tabcolsep}{3.4pt}
\renewcommand{\arraystretch}{1.08}
\begin{tabular}{@{}lccccccc@{}}
\toprule
\textbf{Method} &
\textbf{Turn P} &
\textbf{Turn R} &
\textbf{Turn F1 [95\% CI]} &
\textbf{Turn AP} &
\textbf{Span P} &
\textbf{Span R} &
\textbf{Span F1 [95\% CI]} \\
\midrule
MTAG-Probe &
56.2 & 73.3 &
63.6 [58.4, 69.2] &
92.7 &
98.1 & 85.6 &
91.5 [88.5, 94.3] \\

Target-model self-attribution &
25.8 & 57.1 &
35.6 [31.8, 39.0] &
87.5 &
97.9 & 61.6 &
75.6 [69.8, 81.1] \\

Safety-judge attribution &
30.8 & 55.2 &
39.5 [34.9, 43.8] &
93.1 &
100.0 & 55.7 &
71.5 [65.3, 77.4] \\
\bottomrule
\end{tabular}
\caption{Detailed turn- and span-level localization results.}
\label{tab:localization_detailed}
\footnotesize \raggedright
All methods are evaluated on the same 250 conversations for which both LLM baselines produce valid structured outputs. This cohort contains 100 conversations with positive turn evidence and 150 with no positive turn evidence. Span metrics are computed over the 98 records with scoreable span supervision. Invalid LLM outputs are excluded before metric computation.
\end{table*}

\paragraph{Context-only control.}
When the final user request is protected and deletion is restricted to preceding user turns, MTAG-Probe changes 35.7\% of probe decisions at the 15\% budget, compared with 11.0\% for matched random deletion. In the last-user control (Table~\ref{tab:internal_intervention_detailed}), restricting deletion to the final request changes only 2.2\% of decisions, indicating that most of the intervention effect resides in the preceding conversation. Under the context-only restriction, MTAG-span-random also reaches 35.7\% and LOTO reaches 33.0\%, suggesting that identifying the relevant earlier turns accounts for most of the effect, while the precise span positions within those turns contribute little additional benefit in this setting.

\paragraph{Non-surface control.}
When intervention is restricted to words not matched by the frozen surface-risk lexicon, MTAG-Probe still changes 46.2\% of probe decisions at the 15\% budget, compared with 48.4\% under unrestricted deletion. This indicates that the intervention effect is not primarily driven by overt lexical risk cues.

\begin{table}[!htbp]
\centering
\small
\setlength{\tabcolsep}{3.5pt}
\renewcommand{\arraystretch}{1.08}
\begin{tabular}{@{}lccc@{}}
\toprule
\textbf{Metric} &
\textbf{Before} &
\textbf{After} &
\textbf{$\Delta$} \\
\midrule
Turn F1 &
61.9 & 46.9 &
$-$15.0 [$-$21.5, $-$8.9] \\

Span F1 &
91.4 & 88.0 &
$-$3.4 [$-$6.7, 0.0] \\

False attribution &
26.6 & 27.0 &
+0.4 [$-$5.0, 5.2] \\
\bottomrule
\end{tabular}
\caption{Robustness to semantic user-turn paraphrase.}
\label{tab:paraphrase_results}
\footnotesize
Deltas are paraphrased minus original, with paired cluster-bootstrap
95\% confidence intervals.
\end{table}

\subsection{Semantic Paraphrase Robustness}
\label{app:paraphrase}

We evaluate attribution stability after independently paraphrasing each user turn while preserving assistant turns, conversation structure, and the original evidence labels. Table~\ref{tab:paraphrase_results} compares MTAG-Probe on the original and paraphrased conversations.

Paraphrasing does not significantly increase false attribution on conversations without annotated evidence. The mean Spearman correlation between original and paraphrased turn scores is 0.419, indicating sensitivity in the relative ranking of evidence-bearing turns.

\subsection{Detailed Unsafe-Conversation Detection}
\label{app:detection_details}

Table~\ref{tab:detection_detailed} reports detection performance on the held-out test set and its two constituent corpus types. Performance is stronger on the response-adaptive corpus, whereas the scenario-controlled corpus, authored using frontier-scale models, presents a more challenging detection setting. 

\begin{table}[!htbp]
\centering
\small
\setlength{\tabcolsep}{2.5pt}
\renewcommand{\arraystretch}{1.05}
\begin{tabular}{@{}lrrrrrr@{}}
\toprule
\textbf{Cohort} &
\textbf{$N$} &
\textbf{Prec.} &
\textbf{Rec.} &
\textbf{F1} &
\textbf{FPR} &
\textbf{AUROC} \\
\midrule
All           & 364 & 93.5 & 78.6 & 85.4 & 5.5 & 94.0 \\
Scenario-controlled  & 210 & 88.9 & 68.6 & 77.4 & 8.6 & 89.1 \\
Response-adaptive  & 154 & 98.6 & 92.2 & 95.3 & 1.3 & 97.3 \\
\bottomrule
\end{tabular}
\caption{Detection performance by corpus source.}
\label{tab:detection_detailed}
\end{table}

\subsection{External Detection Generalization}
\label{sec:external_detection}

Safety detection is a supplementary capability of MTAG-Probe. Our primary focus is evidence localization and intervention. We evaluate detection on the three external multi-turn benchmarks. For external detection, all models are evaluated on the conversation prefix ending at the final user turn, excluding the final assistant response.

\begin{table}[!htbp]
\centering
\small
\setlength{\tabcolsep}{1.25pt}
\renewcommand{\arraystretch}{1.05}
\begin{tabular}{@{}lcccc@{}}
\toprule
\textbf{Dataset} &
\textbf{MTAG} &
\textbf{Llama Guard} &
\textbf{ShieldGemma} &
\textbf{$n$}$^\dagger$ \\
\midrule
Geometry & \textbf{45.2} & 11.5 & 12.8 & 8057\\
CoSafe   & \textbf{33.1} & 23.5 & 29.5 & 1400\\
MHJ      & 16.8 & \textbf{35.2} & 28.9 & 537\\
\bottomrule
\end{tabular}
\caption{External detection performance.}
\label{tab:external_detection}
\footnotesize
$^\dagger$ $n$ denotes number of records. Values are detection rates (\%).
\end{table}

Table~\ref{tab:external_detection} shows that MTAG-Probe identifies a larger proportion of source-positive cases than Llama Guard and ShieldGemma on Geometry and CoSafe~\citep{yu-etal-2024-cosafe}. On MHJ~\citep{li2024llm}, MTAG-Probe detects 16.8\% of cases, compared to 35.2\% and 28.9\% for Llama Guard and ShieldGemma, respectively. These results indicate substantial variation in positive detection across external attack distributions. The evaluation does not assess false-positive rates on benign external conversations.

\subsection{Additional Evidence-Localization Results}
\label{app:localization_details}

\paragraph{LLM attribution baselines.}
MTAG-Probe substantially outperforms both LLM attribution baselines. Although safety-judge attribution approaches MTAG-Probe on Top-1 recovery, its substantially higher false attribution rate indicates weaker selectivity over the full conversation.

Target self-attribution queries the source target model (Llama-3-8B-Instruct for response-adaptive and Qwen2.5-32B-Instruct for scenario-controlled conversations), while safety-judge attribution queries the corresponding corpus validator (Mistral-7B-Instruct-v0.3 and Mistral-Small-3.1-24B-Instruct). Localization evaluation uses 354 of the 364 held-out conversations (210 scenario-controlled and 144 response-adaptive). The remaining ten response-adaptive records lack assessed turn labels and are excluded from localization metrics. All 364 conversations remain eligible for detection evaluation. Structured outputs are valid for 301/354 (85.0\%) target-model responses and 258/354 (72.9\%) safety-judge responses. Invalid structured outputs are excluded from all attribution metrics rather than interpreted as empty attribution.

For the primary three-way comparison, we use the intersection of the two valid-output sets, yielding 250 conversations. On this common cohort, MTAG-Probe achieves 63.6 turn F1, compared with 35.6 for target self-attribution and 39.5 for safety-judge attribution. As a sensitivity check using the maximum valid cohort available for each baseline separately, target self-attribution achieves 33.7 turn F1 on 301 conversations compared with 62.8 for MTAG-Probe on the same records, a difference of 29.1 points (95\% CI [23.7, 34.5]). Safety-judge attribution achieves 38.0 turn F1 on 258 conversations compared with 63.5 for MTAG-Probe, a difference of 25.5 points ([20.2, 31.7]). Safety-judge attribution ranks evidence comparably (turn AP 93.1 vs.\ 92.7; Top-1 83.0 vs.\ 85.0), but attributes evidence on 94.7\% of conversations without annotated evidence, so the gap is driven by selectivity rather than ranking.

\begin{table}[!htbp]
\centering
\small
\setlength{\tabcolsep}{2.5pt}
\renewcommand{\arraystretch}{1.08}
\begin{tabular}{@{}lccc@{}}
\toprule
\textbf{Method} &
\textbf{Top-3} &
\textbf{Top-5} &
\textbf{False attribution} \\
\midrule
MTAG-Probe &
95.0 & 97.0 &
28.7 [21.5, 35.6] \\

Target self-attribution &
90.0 & 96.0 &
99.3 [98.0, 100.0] \\

Safety-judge attribution &
95.0 & 97.0 &
94.7 [90.8, 98.0] \\
\bottomrule
\end{tabular}
\caption{Evidence ranking and attribution specificity.}
\label{tab:localization_ranking}
\footnotesize \raggedright
False attribution is the fraction of the 150 no-evidence conversations in the common valid-output cohort receiving any attributed user turn. Confidence intervals are cluster-bootstrap 95\% intervals.
\end{table}

\paragraph{Human validation.}
Human validation is conducted on the 117 conversations judged unsafe by both annotators. Every baseline selects the same number of user turns as MTAG-Probe's top-15\% attribution touches in that conversation. MTAG-Probe intersects Annotator A's essential-turn set in 116/117 cases and Annotator B's in 117/117 cases, yielding a two-human evidence-hit rate of 116/117 (99.1\%; Wilson 95\% CI [95.3, 99.8]). Among the 108 conversations containing jointly agreed essential turns, the annotators identify 126 such turns, of which MTAG-Probe recovers 121 (96.0\% micro recall). The macro-averaged agreed recall reported in Table~\ref{tab:headline_results} is 98.5\%. The corresponding LOTO two-human hit rate is 83/117 (70.9\%), and the paired MTAG-Probe--LOTO comparison gives an exact McNemar $p=2.33\times10^{-10}$. Across ten frozen random repeats, the two-human hit rate averages 34.8\% and ranges from 26.5\% to 41.9\%.

\subsection{Lexical-Risk Specificity}
\label{app:surface_risk}

To examine whether unsafe conversations can be separated using lexical surface cues alone, we evaluate SurfaceRisk as a baseline. Table~\ref{tab:surface_risk_specificity} reports SurfaceRisk sensitivity and specificity across its prespecified thresholds.

\begin{table}[t]
\centering
\small
\setlength{\tabcolsep}{5pt}
\renewcommand{\arraystretch}{1.05}
\begin{tabular}{@{}ccc@{}}
\toprule
\textbf{Threshold} &
\textbf{Benign FPR} &
\textbf{Unsafe TPR} \\
\midrule
0.5 & 50.0 & 59.3 \\
1.0 & 48.9 & 58.8 \\
1.5 & 43.4 & 51.1 \\
2.0 & 37.4 & 46.2 \\
2.5 & 27.5 & 38.5 \\
3.0 & 4.9  & 18.1 \\
\bottomrule
\end{tabular}
\caption{SurfaceRisk lexical specificity.}
\label{tab:surface_risk_specificity}
\footnotesize
FPR is the fraction of benign conversations exceeding the lexical-risk
threshold; TPR is the corresponding fraction of unsafe conversations.
\end{table}

SurfaceRisk frequently misclassifies benign conversations as unsafe and exhibits a substantial specificity--sensitivity trade-off. Thresholds that retain approximately half of unsafe conversations misclassify a large fraction of benign conversations, while reducing the benign false-positive rate to 4.9\% lowers unsafe coverage to 18.1\%. The effect is stronger for response-adaptive benign conversations, which retain substantial lexical overlap with their adversarial counterparts. At threshold 0.5, SurfaceRisk fires on 84.4\% of response-adaptive benign conversations, compared with 24.8\% of scenario-controlled benign conversations.

\begin{table*}[t]
\centering
\small
\setlength{\tabcolsep}{4.0pt}
\renewcommand{\arraystretch}{1.08}
\begin{tabular}{@{}llrrrccc@{}}
\toprule
\textbf{Dataset} &
\textbf{Safety guard} &
\textbf{Budget} &
\textbf{$N$} &
\textbf{$n$} &
\textbf{MTAG-Probe flip} &
\textbf{$\Delta$ vs.\ Random} &
\textbf{$\Delta$ vs.\ SurfaceRisk} \\
\midrule
Geometry &
Llama Guard &
15\% &
8,057 &
926 &
50.6 [44.4, 57.8] &
+33.0 [26.4, 40.2] &
+43.4 [35.3, 52.6] \\

Geometry &
Llama Guard &
20\% &
8,057 &
926 &
60.8 [53.9, 68.0] &
+37.5 [30.3, 44.7] &
+54.1 [45.5, 63.0] \\

Geometry &
ShieldGemma &
15\% &
8,057 &
1,029 &
28.7 [23.7, 35.4] &
+13.5 [10.2, 17.2] &
+21.7 [16.1, 28.6] \\

Geometry &
ShieldGemma &
20\% &
8,057 &
1,029 &
34.7 [28.5, 43.3] &
+18.0 [14.7, 22.2] &
+25.6 [18.6, 33.8] \\
\bottomrule
\end{tabular}

\caption{External safety-guard transfer on the full harmful Geometry cohort.}
\label{tab:external_transfer_detailed}

\footnotesize \raggedright
$N$ denotes the full source-positive Geometry cohort.
$n$ denotes conversations judged unsafe before intervention and eligible for guard-specific flip evaluation.
Confidence intervals are cluster-bootstrap 95\% intervals.
Paired differences are MTAG-Probe minus the indicated equal-budget baseline.
\end{table*}

\subsection{Hard-Benign Stress Evaluation}
\label{app:hard_benign}

Table~\ref{tab:hard_benign_specificity} reports false attribution on this
stress set. A conversation counts as a false attribution for MTAG-Probe when
at least one user turn exceeds the frozen attribution threshold. For the two
LLM baselines, a false attribution is a nonempty returned evidence-turn set.
SurfaceRisk fires when at least one user word reaches its frozen lexical-risk
threshold.

\begin{table}[!htbp]
\centering
\small
\setlength{\tabcolsep}{4pt}
\renewcommand{\arraystretch}{1.08}
\begin{tabular}{@{}lrrr@{}}
\toprule
\textbf{Method} &
\textbf{False attr.} &
\textbf{Rate} &
\textbf{95\% CI} \\
\midrule
MTAG-Probe
    & 248/587 & 42.2 & [36.2, 48.1] \\
SurfaceRisk
    & 218/587 & \textbf{37.1} & -- \\
Target self-attribution
    & 551/587 & 93.9 & [91.6, 95.9] \\
Safety-judge attribution
    & 479/587 & 81.6 & [77.5, 85.4] \\
\bottomrule
\end{tabular}
\caption{Specificity on hard-benign conversations.}
\label{tab:hard_benign_specificity}
\footnotesize
Lower is better. Confidence intervals use 10,000 scenario-family bootstrap replicates where available.
\end{table}

MTAG-Probe is substantially more selective than either LLM attribution
baseline on behaviorally benign trajectories. SurfaceRisk is slightly more
conservative at the conversation level, consistent with its role as a sparse
lexical heuristic. MTAG-Probe attributes 731 of 3,683 individual benign user
turns (19.8\%), showing that the conversation-level rate partly reflects the
accumulation of attribution opportunities across multi-turn interactions.

The trajectory-level detector produces 120 false positives among the 587
conversations, corresponding to a 20.4\% false-positive rate
(95\% CI: 15.8--25.3). This stress evaluation is substantially more demanding
than ordinary benign evaluation because every retained trajectory was
specifically constructed as a difficult benign control and subsequently
validated on the realized target-model behavior.

Both LLM attribution baselines produce valid structured outputs for all 587 stress conversations, so no records are excluded from these results.

\subsection{External Safety-Guard Transfer}
\label{app:external_transfer}

External attribution is evaluated on the full source positive subset selected before MTAG-Probe inference. Geometry contains 16,016 prepared conversations, including 8,057 harmful/source-positive conversations.

For each external safety guard, unsafe$\rightarrow$safe intervention effects are then computed only on candidate conversations that the guard labels unsafe before deletion. This yields 926 conversations for Llama Guard and 1029 conversations for ShieldGemma. Candidate selection is independent of MTAG-Probe outputs. Table~\ref{tab:external_transfer_detailed} reports the resulting unsafe$\rightarrow$safe transition rates and paired differences at the 15\% and 20\% user-word deletion budget. Both paired MTAG-Probe improvements exclude zero under both external guards. 

\paragraph{Trajectory-length sensitivity.}
The unrestricted external evaluation does not impose a minimum conversation length. As a post-hoc sensitivity analysis, we separate trajectories with 1--5 user turns from those with at least six user turns. Table~\ref{tab:external_length} reports the corresponding MTAG-Probe intervention effects. MTAG-Probe does not show any consistent monotonic dependence on the trajectory length.

\begin{table}[!htbp]
\centering
\small
\setlength{\tabcolsep}{3.0pt}
\renewcommand{\arraystretch}{1.05}
\begin{tabular}{@{}llrrrr@{}}
\toprule
&
&
\multicolumn{2}{c}{\textbf{1--5 user turns}} &
\multicolumn{2}{c}{\textbf{$\geq$6 user turns}} \\
\cmidrule(lr){3-4}\cmidrule(l){5-6}
\textbf{Guard} &
\textbf{Budget} &
\textbf{$n$} &
\textbf{Flip} &
\textbf{$n$} &
\textbf{Flip} \\
\midrule
Llama Guard &
15\% &
55 &
38.2 &
871 &
51.4 \\

Llama Guard &
20\% &
55 &
41.8 &
871 &
62.0 \\

ShieldGemma &
15\% &
435 &
32.2 &
594 &
26.1 \\

ShieldGemma &
20\% &
435 &
38.9 &
594 &
31.6 \\
\bottomrule
\end{tabular}
\caption{External safety-guard transfer on Geometry by trajectory length.}
\label{tab:external_length}
\footnotesize
$n$ denotes conversations judged unsafe before intervention. Flip values are unsafe$\rightarrow$safe transition
rates (\%).
\end{table}

\subsection{Target-Model Gradient Attribution}
As a supplementary comparison, we evaluate two standard white-box attribution methods directly on the source-specific target model. Integrated Gradients uses 32 midpoint steps with a zero-embedding baseline over user-content tokens, while Grad$\times$Input uses the gradient at the observed input. Both methods attribute the mean teacher-forced log-likelihood of the first observed unsafe assistant response to preceding user evidence and retain positive signed contributions for ranking. We use Llama-3-8B for response-adaptive trajectories and Qwen2.5-32B for scenario-controlled trajectories. Because these methods require a stored unsafe target-model response, the comparison is restricted to the 146 adversarial test trajectories for which such a response is available and to user turns preceding that response.

MTAG-Probe performs better than both gradient baselines across turn- and span-level ranking metrics (Table~\ref{tab:gradient-baselines}), exceeding the stronger turn-level baseline, Grad$\times$Input, by 8.4 points in Turn AP and 11.9 points in Turn Top-1. Turn AP uses 86 trajectories with both positive and negative assessed turns, while Turn Top-1 uses 109 trajectories with positive evidence before the target response. Span AP uses 15,203 commonly scoreable characters from 99 trajectories. Character-level scoring provides a common unit across the target-model and MTAG-Probe tokenizations.

Because 82.4\% of scoreable characters are positive, the random-ranking baseline for Span AP is approximately 82.4. Integrated Gradients performs near this level and Grad$\times$Input only 2.6 points above it, whereas MTAG-Probe reaches 98.8. Span AUROC further separates the methods, reaching 94.4 for MTAG-Probe compared with 53.3 for Integrated Gradients and 51.7 for Grad$\times$Input. Gradients measure local sensitivity of the likelihood, while benchmark evidence is defined by discrete deletion and regeneration, so the two can diverge when an effect is distributed across a span or saturates.

\begin{table}[t]
\centering
\small
\setlength{\tabcolsep}{4.5pt}
\renewcommand{\arraystretch}{1.05}
\begin{tabular}{lrrr}
\toprule
\textbf{Method} & \textbf{Turn AP} & \textbf{Span AP} & \textbf{Turn Top-1} \\
\midrule
\textbf{MTAG-Probe}  & \textbf{91.5} & \textbf{98.8} & \textbf{83.5} \\
Grad$\times$Input    & 83.1 & 85.0 & 71.6 \\
Integrated Gradients & 76.0 & 82.5 & 49.5 \\
\bottomrule
\end{tabular}
\caption{Target-model gradient attribution baselines on the target-supported adversarial test cohort.}
\label{tab:gradient-baselines}
\end{table}

\subsection{Ablation Study}
\label{app:ablations}

We evaluate four variants of the full MTAG-Probe model using the same held-out test set and a 15\% intervention budget. \textsc{No Cross-Turn} removes the cross-turn Transformer, making each turn representation independent of other turns before conversation-level aggregation. \textsc{Frozen Backbone} freezes all ModernBERT layers. \textsc{Single-Stage} removes the detection-only warm-up and localization-loss ramp, enabling all objectives from the beginning of training. \textsc{Detection-Only} removes explicit turn- and span-localization losses and selects the checkpoint using detection performance. Its attributions are read from the untrained localization heads. All results use seed 42.

\end{document}